\documentclass[sigconf]{acmart}
\usepackage{url}

\AtBeginDocument{%
  \providecommand\BibTeX{{%
    \normalfont B\kern-0.5em{\scshape i\kern-0.25em b}\kern-0.8em\TeX}}}

\renewcommand\footnotetextcopyrightpermission[1]{}
\setcopyright{none}
\acmDOI{}

\begin{document}
\fancyhead{}

\title{Mesh Guided One-shot Face Reenactment Using Graph
	Convolutional Networks}

\author{Guangming Yao}
\authornote{Both authors contributed equally to this research.}
\email{yaoguangming@corp.netease.com}
\affiliation{%
  \institution{NetEase Fuxi AI Lab}}

\author{Yi Yuan}
\authornotemark[1]
\email{yuanyi@corp.netease.com}
\affiliation{%
\institution{NetEase Fuxi AI Lab}}
 
\author{Tianjia Shao}
\authornote{Corresponding author}
\email{tjshao@zju.edu.cn}
\affiliation{
   \institution{State Key Lab of CAD\&CG, Zhejiang University}}

\author{Kun Zhou}
\email{kunzhou@acm.org}
 \affiliation{
   \institution{State Key Lab of CAD\&CG, Zhejiang University}}
   
\begin{abstract}
	Face reenactment aims to animate a source face image to a different pose and expression provided by a driving image.Existing approaches are either designed for a specific identity, or suffer from the \textit{identity preservation} problem in the one-shot or few-shot scenarios.
	In this paper, we introduce a method for one-shot face reenactment, which uses the reconstructed 3D meshes (i.e., the source mesh and driving mesh) as guidance to learn the optical flow needed for the reenacted face synthesis. Technically, we explicitly exclude the driving face's identity information in the reconstructed driving mesh. In this way, our network can focus on the motion estimation for the source face without the interference of driving face shape. We propose a motion net to learn the face motion, which is an asymmetric autoencoder. The encoder is a graph convolutional network (GCN) that learns a latent motion vector from the meshes, and the decoder serves to produce an optical flow image from the latent vector with CNNs. Compared to previous methods using sparse keypoints to guide the optical flow learning, our motion net learns the optical flow directly from 3D dense meshes, which provide the detailed shape and pose information for the optical flow, so it can achieve more accurate expression and pose on the reenacted face.
	Extensive experiments show that our method can generate high-quality results and outperforms state-of-the-art methods in both qualitative and quantitative comparisons.
\end{abstract}



\begin{teaserfigure}
\centering
\scalebox{0.21}{\includegraphics{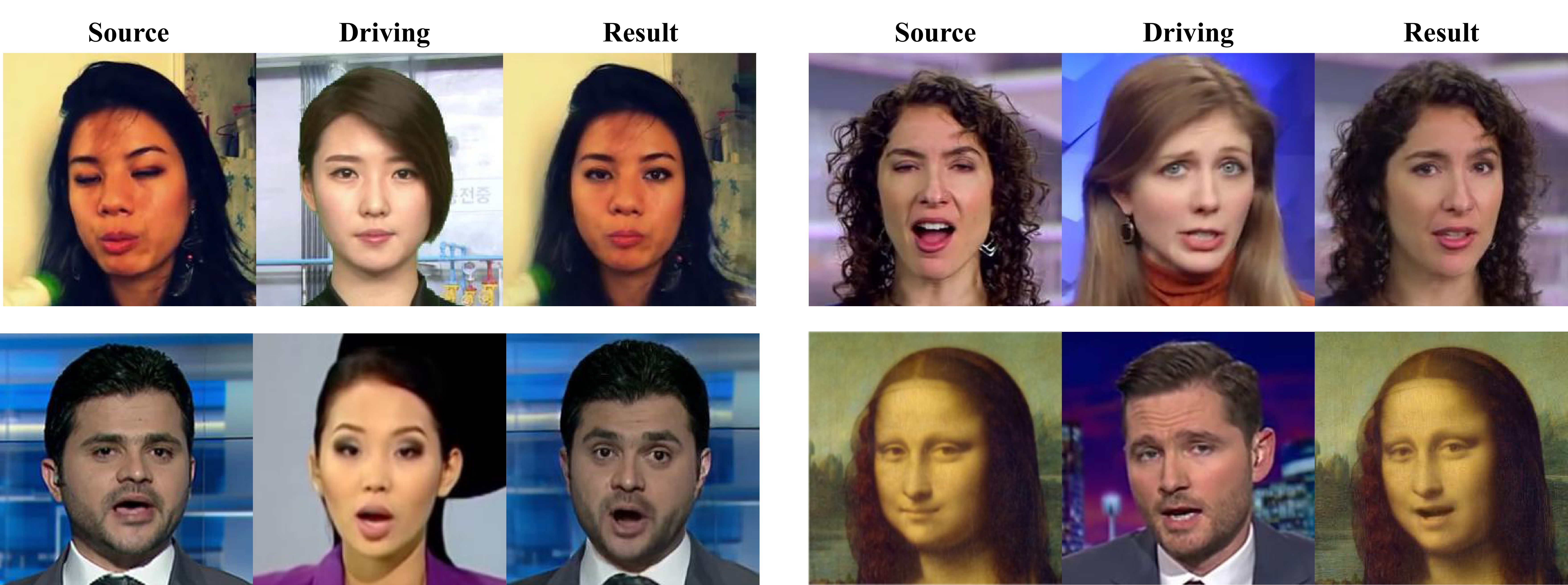}}
 \caption{Examples generated by our method.}
 \Description{}
 \label{fig:teaser}
\end{teaserfigure}

\maketitle

\fancyfoot{}
\thispagestyle{empty}

\section{Introduction}

In this paper, we propose a mesh guided one-shot face reenactment framework, which animates a source face image using a driving face image of a different person, as shown in Fig.~\ref{fig:teaser}. The reenacted face preserves the same identity as the source person, while owning the pose and expression of the driving person.

Previous approaches have demonstrated great success in face reenactment for a specific identity using generative adversarial networks (GANs). For instance, the work of ~\cite{wu2018reenactgan,xu2017face} successfully performs face reenactment between a pair of specified identities using CycleGAN~\cite{zhu2017unpaired}. High-fidelity results are achieved with the help of 3D face reconstruction and GANs~\cite{thies2016face2face,kim2018deep}. Nevertheless, all these methods require a large number of images for a specific identity, which may be infeasible for many applications. Consequently, a variety of one-shot or few-shot methods are proposed~\cite{zakharov2019fewshot,wiles2018x2face,Siarohin_2019_CVPR,Siarohin_2019_NeurIPS}.

The state-of-the-art one-shot or few-shot methods utilize the optical flow to estimate the face motion from the source image to the driving image. The optical flow is learned from scratch under the guidance of sparse keypoints~\cite{Siarohin_2019_CVPR,Siarohin_2019_NeurIPS}. However, there are two limitations with such strategy. First, as the optical flow is extracted from the two images, the shape of driving face is inevitably involved in the optical flow estimation. Due to the interference of driving face shape, such optical flow cannot purely reflect the motion of source face. As a result, the source identity may not be well preserved after feature warping using such optical flow (see Fig.~\ref{fig:cmp} for example). Although the latest few-shot work of~\cite{ha2019marionette} can alleviate this problem by introducing the \textit{Landmark Transformer} mechanism, the result is still not satisfactory for the one-shot case. Second, the sparse keypoints used to guide the optical flow learning cannot faithfully represent the full face expression and pose. Consequently, the expression and pose of the reenacted face may not match well with the driving face (see Fig.~\ref{fig:relateive_cmp} for example).

To tackle these problems, we propose a novel one-shot face reenactment framework which deploys the reconstructed 3D meshes as the 3D dense shape guidance for optical flow estimation. 
First, to account for the source identity preservation, in the mesh regression module (Section~\ref{sec:3.1}), we explicitly exclude the driving face's identity information in the reconstructed driving face mesh. That is, while the source mesh is reconstructed with all the regressed source parameters (i.e. identity, expression and pose), we build the driving mesh with the source identity and driving pose and expression. In this way, we can focus on the motion estimation for the source face without the interference of driving face shape, and the source identity can be well preserved (see Fig.~\ref{fig:cmp} for example). The meshes are then transformed with the pose to match the face images, so that they can serve as the guidance for the image optical flow estimation. 
Next, to obtain an optical flow image aware of the full face expression and pose, in a key stage, we design a motion net (Section~\ref{sec:3.2}) to predict the optical flow image from the dense meshes, which can provide the detailed shape and pose information for the optical flow. Specifically, the motion net is designed as an asymmetric autoencoder, where the encoder is a graph convolutional network (GCN) serving to extract motion features from the two meshes, and the decoder is 2D convolution based image decoder for estimating the optical flow image. With such optical flow estimated from dense meshes, our framework is able to produce a reenacted face whose expression and pose closely match the driving face (see Fig.\ref{fig:reenacting} for example). At last, we send the source image and the well estimated optical flow to the reenacting module, which utilizes the feature warping strategy~\cite{grigorev2018coordinate,siarohin2018deformable,Siarohin_2019_CVPR,Siarohin_2019_NeurIPS,ren2020deep} to fuse the source appearance and motions to produce the reenacted image (Section~\ref{sec:3.3}). The motion net and the reenacting module are trained together in an end-to-end way.

We qualitatively and quantitatively evaluated our method on different datasets. Experimental results show that our one-shot method is able to better preserve the identity and yield a reenacted face with more accurate expression and pose, using only one source image and one driving image that are both unseen in the training data. Our method outperforms the state-of-the-art methods in both objective and subjective aspects.

The main contributions of our work are:
\begin{itemize}
	\item We propose a novel mesh-guided one-shot method for face reenactment, which explicitly excludes the interference of driving face shape for source identity preservation, and estimates optical flows from dense meshes to obtain the accurate pose and expression for the reenacted face.
	
	\item To the best of our knowledge, we are the first to use GCNs to directly learn the motion from meshes for face reenactment. The estimated optical flow is aware of the detailed shape and pose information from the meshes.
	
	\item We compare our results with state-of-the-art methods, and ours outperform others in both qualitative and quantitative comparison.
\end{itemize}

\section{Related Work}

\begin{figure*}[ht]
	\centering
	\scalebox{0.9}{\includegraphics[width=\linewidth]{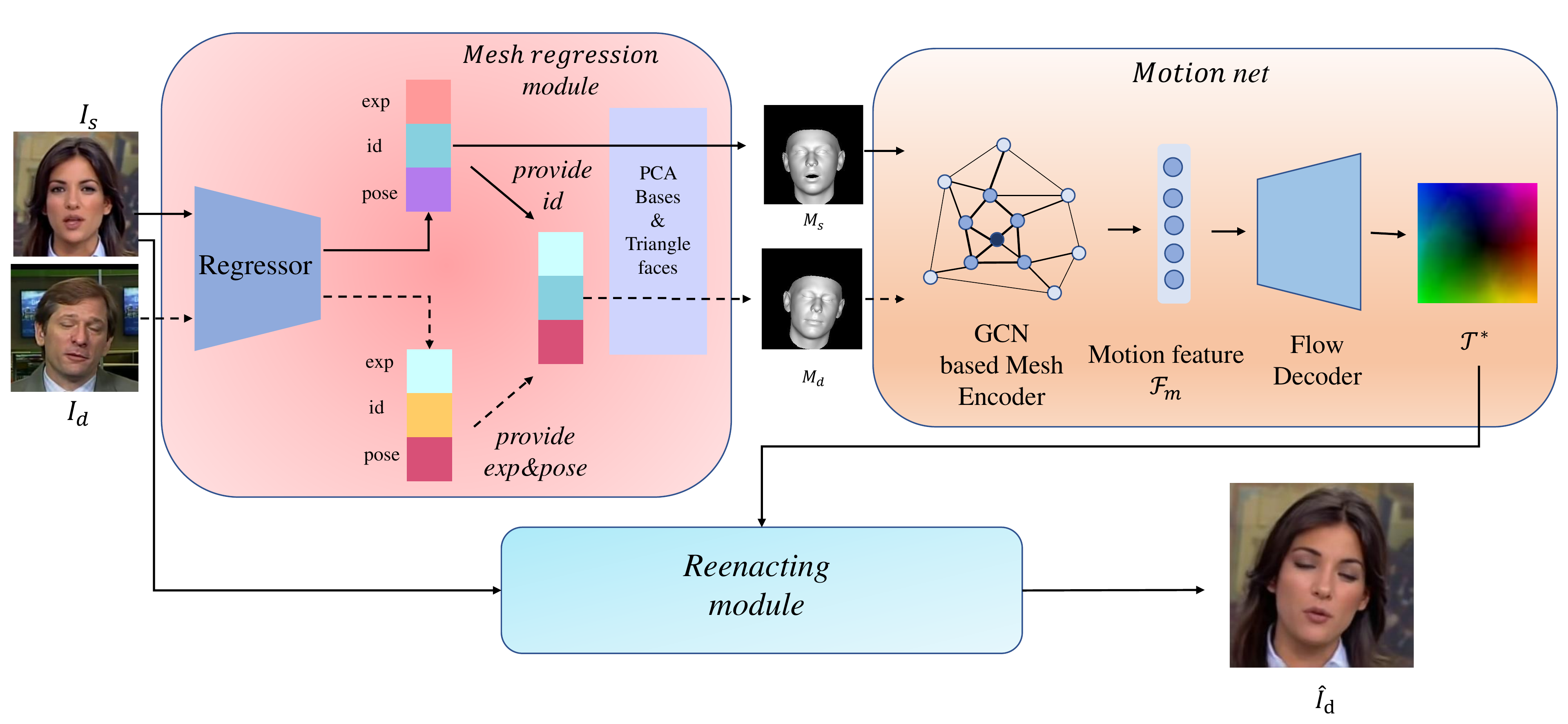}}
	\caption{The generator of the proposed approach. The regressor predicts the 3DMM parameters (i.e., identity, expression and pose) from both the source image $I_s$ and the driving image $I_d$, so as to reconstruct the source mesh $M_s$ and driving mesh $M_d$.
	$M_s$ and $M_d$ are stacked and fed to the motion net to produce the optical flow ${\mathcal{T^*}}$. ${\mathcal{T^*}}$ is then sent to the reenacting module to output the reenacted image ${\hat{I}_d}$.
	}
	\Description{The main method.}
	\label{fig:method}
\end{figure*}

\subsection{3DMM based 3D Face Reconstruction}

Since 3D Morphable Model(3DMM)~\cite{blanz1999morphable} was proposed in 1999, there have been many variations of 3DMM ~\cite{booth2018large,cao2013facewarehouse,huber2016multiresolution,gerig2018morphable,li2017learning} for single-image based 3D face reconstruction. 
These models produce low-dimensional representations for the face
identity, expression and texture from multiple face scans using PCA.
One of the most widely used, publicly available variants of 3DMM is the Basel Face Model (BFM)~\cite{paysan20093d}. We use the BFM as our 3DMM model in this paper for generating 3D face meshes.
Recently, deep learning based methods directly regress the 3DMM coefficients from images either in a supervised way~\cite{dou2017end,zhu2017face,jourabloo2015pose,tuan2017regressing,lin2020highfidelity} or in an unsupervised way~\cite{tewari2017mofa,bas20173d}.
Our method adopts ~\cite{zhu2017face} to regress the identity, pose and expression parameters for the source image and driving image respectively.

\subsection{Face Reenactment}
The face reenactment aims to animate a source image to another pose-and-expression, which is provided by the driving image.
Recent works demonstrate great success in face reenactment for a specified identity. For instance, 
ReenactGAN~\cite{wu2018reenactgan} drives a specific identity with the encoder-decoder framework.
Face2Face~\cite{thies2016face2face} animates the facial expression of source video by utilizing the rendered image. 
The work of~\cite{kim2018deep} transfers the full 3D head position, head rotation, face expression, eye gaze, and eye blinking from a driving actor to a portrait video of source actor.
However, all these methods require a large number of images of the specific identity for training, and only reenact the face of the specific identity.
In contrast, our method does not have this limitation, and is capable of reenacting any identity given only a single image without the need for retraining or fine-turning. 

To extend face reenactment to unseen identities, some one-shot or few-shot methods\cite{zakharov2019fewshot,geng2018warp,wiles2018x2face,Siarohin_2019_CVPR,Siarohin_2019_NeurIPS,ha2019marionette} have been proposed.
Recently, ~\cite{zakharov2019fewshot} adopts the meta-learning mechanism, which is able to synthesize a personalized talking head with several images.
But it requires fine-tuning when a new-coming person is introduced.
~\citet{zhang2019oneshot} introduce a unsupervised approach for face reenactment, which needs no different pose for the same identity. Yet, the face parsing map, a vary identity-specific feature, is utilized to guide the reenacting, which leads to distorted results when reenacting a different identity.
~\citet{geng2018warp} introduce warp-guided GANs for single-photo facial animation. However, their method needs a photo with frontal pose and neutral expression, while ours does not have this limitation.
X2Face~\cite{wiles2018x2face} proposes an approach to effectively utilize multiple source images, but it cannot synthesize face regions which do not exist in source images.
The methods in~\cite{Siarohin_2019_CVPR,Siarohin_2019_NeurIPS} can learn motion for universal objects and perform the relative motion transfer. 
However, the initial driving image with source pose-and-expression is required to correctly reenacted image, which is hard to be fulfilled in many applications.
More notably, the main problem in one-shot or few-shot face reenactment is the \textit{identity preservation} problem, that is, the source face shape is difficult to be preserved during reenacting when there is a large shape difference between the source and driving faces.
Although the latest few-shot work of~\cite{ha2019marionette} can alleviate this problem by introducing the \textit{Landmark Transformer} mechanism, the result is still not satisfactory for the one-shot cases.
In our approach, we explicitly exclude the driving face's identity information in the reconstructed driving face mesh. Thus our network can focus on learning the motion for the source face without the interference of driving face shape.

\subsection{Graph Convolutional Network}

To apply convolutional operations to non-Euclidean structured data, ~\citet{bruna2013spectral} adopt the graph Laplacian and Fourier basis to enable the first extension of CNNs on graphs. 
Since then, there have been increasing improvements, extensions, and approximations on GCNs~\cite{henaff2015deep,defferrard2016convolutional,kipf2016semi,levie2018cayleynets}. 
Recently, CoMA~\cite{ranjan2018generating} introduces mesh down-sampling and mesh up-sampling layers, and constructs an autoencoder to learn a latent embedding of 3D face meshes.
Inspired by CoMA, our method establishes an asymmetric autoencoder which employs GCNs to learn a latent vector representing the relative motion from the source mesh to the driving mesh.

\section{Approach}

Our mesh-guided face reenactment network consists of a generator and a discriminator. While the discriminator is deployed from WGAN-GP~\cite{gulrajani2017improved}, the core of our method is the generator. As shown in Fig.~\ref{fig:method}, the generator is composed of three modules.
The mesh regression module is utilized to regress the 3DMM coefficients and face pose from the source image $I_s$ and driving image $I_d$ respectively (Section ~\ref{sec:3.1}). The source mesh $ M_s $ and driving mesh $ M_d $ are constructed with the estimated coefficients, and transformed using the pose to match the face images. In this way, we can establish explicit 3D dense guidance for the face images to learn a good initial optical flow between the images from the aligned meshes. Given $M_s$ and $M_d$ as input, the motion net is deployed to estimate such optical flow (Section~\ref{sec:3.2}). The motion net is designed with an asymmetric autoencoder architecture. The encoder is a GCN-based mesh encoder, which performs graph convolutional operations on the meshes to extract a motion feature $\mathcal{F}_m$ (as shown in Fig.~\ref{fig:method}). The decoder is a 2D convolution based image decoder, which performs up-sampling convolutional operations to produce an optical flow image. Afterwards, we send the well estimated optical flow and source image to the reenacting module, which fuses the source appearance features and occlusion-aware motions~\cite{Siarohin_2019_NeurIPS} to produce the reenacted face image  (Section \ref{sec:3.3}). 
We adopt a pre-trained 3DMM regressor~\cite{zhu2017face} in the mesh regression module. The motion net and reenacting module are jointly trained in an end-to-end way. To ensure realistic results, we adopt adversarial training using the discriminator from ~\cite{gulrajani2017improved}.

\subsection{Mesh Regression Module}\label{sec:3.1}

The mesh regression module constructs the source and driving meshes needed for motion estimation. For this task, we adopt 3DMM regressor~\cite{zhu2017face} to regress the 3DMM coefficients and face pose from the input images with CNNs. 
Given a 2D image, it regresses a vector $c = (c_i,c_e,p) \in \mathbb{R}^{113}$, where $c_i\in \mathbb{R}^{50}$ and $c_e\in \mathbb{R}^{51}$ represent the 3DMM identity and expression coefficients respectively. $p\in \mathbb{R}^{12}$ is the face pose. With the predicted coefficients, the 3D coordinates $V$ of face vertices can be computed with:
\begin{equation}
V = V_{mean} + c_i V_{id} + c_e V_{exp},
\end{equation}
where $V_{mean}$ and $V_{id}$ are PCA bases from BFM~\cite{paysan20093d}, and $V_{exp}$ is built from~\cite{cao2013facewarehouse}. The mesh topology is adopted from BFM~\cite{paysan20093d}, which is the same for all estimated meshes. The meshes are then transformed with the face pose to match the face images, so that they can serve as the 3D dense guidance for subsequent optical flow estimation between images.

Fig.~\ref{fig:method} illustrates the procedure of our mesh regression module. It is worth noting that, while the source mesh $M_s$ is built with the parameters from the source image $I_s$, the driving mesh $M_d$ is constructed with a combination of parameters, that is, the source identity from $I_s$ and the driving expression and face pose from the driving image $I_d$. By doing so, we can estimate the facial movements for the source person without the interference of the driving person's face shape.

\subsection{Motion Net}\label{sec:3.2}

With the reconstructed meshes as the explicit 3D dense guidance, the task of learning the face motion from $ I_s $ to $ I_d $ can be largely facilitated. Specifically, we design an asymmetric autoencoder called motion net to learn an optical flow image from the two meshes. Given $M_s = (\mathcal{V}_s, \mathcal{A})$ and $M_d = (\mathcal{V}_d, \mathcal{A})$, where $\mathcal{V}_s, \mathcal{V}_d \in \mathbb{R}^{n\times3}$ store vertex coordinates and $\mathcal{A} \in \{0,1\}^{n\times n}$ is the adjacency matrix representing the connectivity between vertices, we stack $\mathcal{V}_s$ and $\mathcal{V}_d$ as $\mathcal{V}^* \in \mathbb{R}^{n\times6}$ to obtain a stacked mesh $M^*= (\mathcal{V}^*, \mathcal{A})$. The motion net takes as input the stacked mesh $M^*$, and outputs an estimated optical flow $\mathcal{T^*} \in \mathbb{R}^{H\times W \times 2}$ from $I_s$ to $I_d$, where $H,W$ are the image height and width.

Technically, the encoder is a graph convolutional network to extract the motion feature vector from the stacked mesh. To perform graph convolution on the mesh, we calculate the normalized graph Laplacian matrix as $\mathcal{L} = \mathcal{I} - \mathcal{D}^{-\frac{1}{2}} \mathcal{A} \mathcal{D}^{-\frac{1}{2}}$, where $\mathcal{I} \in \mathbb{R}^{n\times n}$ is the identity matrix, and $ \mathcal{D} \in \mathbb{R}^{n\times n}$ is  diagonal matrix of vertex degrees, i.e., $ \mathcal{D}_{ii} = \sum_j (\mathcal{A}_{ij})$. 
Chebyshev polynomials~\cite{defferrard2016convolutional} is adopted by our method, and the spectral convolution is then defined as:
\begin{equation}
g_\theta(\mathcal{L}) = \sum_{k=0}^{K-1}\theta_k T_k (\widetilde{\mathcal{L}}),
\end{equation}
\begin{equation}
y_j = \sum_{i=1}^{C_{in}} g_{\theta_{ij}}(\mathcal{L})x_i,
\end{equation}
where $ \widetilde{\mathcal{L}} = 2\mathcal{L} / \lambda_{max} - \mathcal{I} $  is the scaled Laplacian matrix, $ \lambda_{max} $ is the maximum eigenvalue of the normalized Laplacian matrix, $ \theta_k $ is the Chebyshev coefficients vector, and $ T_k \in  \mathbb{R}^{n \times n}  $ is the Chebyshev polynomial of order $ k $. $T_k$ is computed recursively as
$ T_k(x) = 2xT_{k-1}(x) - T_{k-2}(x) $ with the initial $ T_0 = 1 $ and $ T_1 = x $. $ x_i \in \mathbb{R}^n $ is the $i$ th channel of input $ x \in \mathbb{R}^{n\times C_{in}}$, and $ y_j \in \mathbb{R}^n $ denotes the $ j $ th channel of output $ y \in \mathbb{R}^{n\times C_{out}} $. 

The encoder architecture is built following the idea of residual networks, which consists of four spectral residual
blocks~\cite{ranjan2018generating}. The graph residual connections~\cite{kolotouros2019cmr} are deployed as they help speed up the training process and also improve the output quality. A mesh down-sampling layer~\cite{ranjan2018generating} is placed between two residual blocks to integrate information among neighboring vertices. 
Each residual block contains one Chebyshev convolutional layer and one graph linear layer~\cite{kolotouros2019cmr}. Every Chebyshev convolutional layer uses Chebyshev polynomials K = 3  and is followed by  a instance normlization\cite{ulyanov2016instance} and a ReLU layer~\cite{glorot2011deep}.

The decoder is an image decoder using 2D convolutions. It contains several 2D convolutional up-sampling blocks to expand the spatial size of feature maps, as well as several residual blocks~\cite{brock2018large} to refine the result. There are two branches in the up-sampling block. The main branch has an up-sampling layer and two convolutional layers, and the short cut branch has only an up-sampling layer. The outputs of two branches are element-wisely added as the block output. Each layer in the up-sampling block is followed by a BN layer~\cite{ioffe2015batch} and a ReLU layer~\cite{glorot2011deep}.
Please refer to the supplementary material for more details.

\subsection{Reenacting Module}\label{sec:3.3}
\begin{figure}[h]
	\centering
	\includegraphics[width=\linewidth]{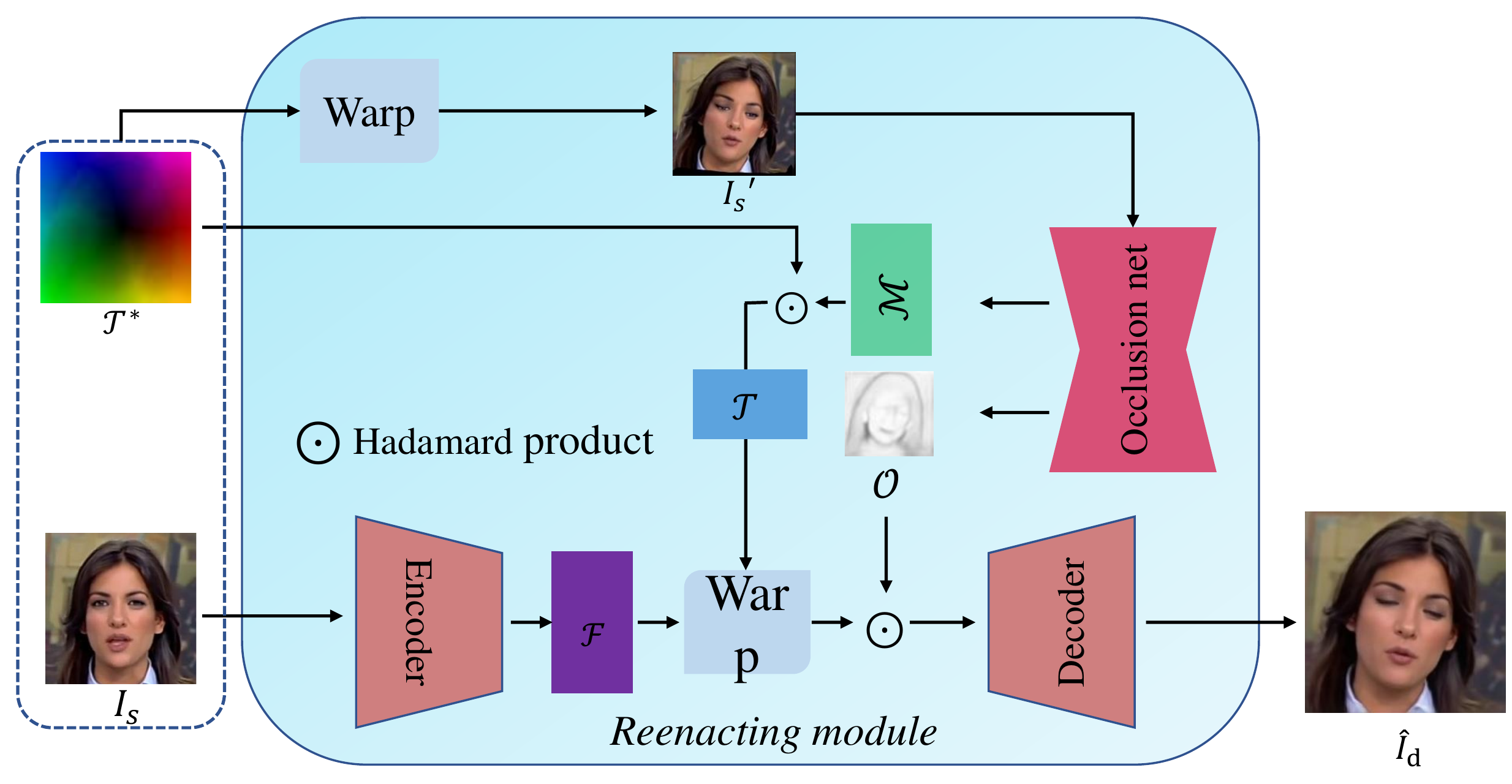}
	\caption{The reenacting module. The source image ${I_s}$ and estimated optical flow ${\mathcal{T^*}}$ are fed into the reenacting module to produce the reenacted image ${\hat{I}_d}$. The occlusion net here is utilized to generate the occlusion map $\mathcal{O}$ and the optical flow mask $M$, which are used to diminish the impact of the features corresponding to the occluded parts.}
	\Description{The main method.}
	\label{fig:reenacting}
\end{figure}

The architecture of our reenacting module is similar to those of state-of-the-art works~\cite{Siarohin_2019_CVPR,Siarohin_2019_NeurIPS}, which adopts the feature warping strategy to avoid the pixel-misalignment between the source and target images.
However, a key difference between ours and previous works is that our module does not require the part of learning the optical flow from scratch. Thanks to the meshes as explicit 3D dense guidance, our motion net can provide a well estimated optical flow to our reenacting module.

As shown in Fig.~\ref{fig:reenacting}, the reenacting module takes as input the estimated optical flow $\mathcal{T^*}$ and the source image $I_s$, and it contains an encoder, an occlusion net and a decoder. The encoder serves to extract the appearance features from the $I_s$. Similar to ~\cite{Siarohin_2019_NeurIPS}, the occlusion net learns to produce an occlusion map and an optical flow mask from the warped source image $I_s'$ with $\mathcal{T^*}$. The occlusion map and optical flow mask take occlusion into account and assign different confidence values for the estimated movements on pixels. The encoded appearance feature maps are warped using the masked optical flow. Then they are fused with the occlusion map and sent to the decoder. In this way, the decoder can be aware of both the source appearance and motions with different confidences, and learns to produce the reenacted face image.
The occlusion net is a four-layer hourglass net~\cite{yang2017stacked}. The encoder and decoder are composed of several convolutional down-sampling and up-sampling blocks respectively. For more details, please refer to the supplementary material.

\subsection{Loss Functions}

Following previous works~\cite{ha2019marionette,Siarohin_2019_CVPR,Siarohin_2019_NeurIPS}, we adopt the self-supervised approach to jointly train the motion net and reenacting module. The identities of source image and driving image are same in training stage, but are different in inference stage.
The loss of our method is defined by:
\begin{equation}
\begin{aligned}
L_{total} (I_s, I_d,\hat{I}_d) &= \lambda_{rec} L_{rec}(I_d,\hat{I}_d) + \lambda_{c} L_{c}(I_s, I_d,\hat{I}_d)\\
&+\lambda_{FM} L_{FM}(I_d,\hat{I}_d) + \lambda_{adv} L_{adv}(I_d,\hat{I}_d),
\end{aligned}
\end{equation}
where $L_{rec}$ denotes the reconstruction loss, $L_{c}$ denotes the coefficient loss, $L_{FM}$ denotes the feature matching loss and $L_{adv}$ denotes the adversarial loss. 

\subsubsection{Reconstruction loss}
A straightforward objective is to minimize the difference between the driving images and the self-reenacted images. We utilize the pre-trained VGG~\cite{parkhi2015deep} to calculate the perceptual loss as the reconstruction loss. Following~\cite{Siarohin_2019_NeurIPS}, the reconstruction loss is computed by accumulating the perceptual losses calculated in different resolutions to reduce blurriness:
\begin{equation}
L_{rec} (I_d,\hat{I}_d) =  \sum_{p=1}^P \sum_{i=1}^L ||F_i(\hat{I}_d) - F_i(I_d)|| ,
\end{equation}
where $ F_i (\cdot)$ denotes the $ i^{th} $ feature extracted from a VGG layer and $ L $ is the number of features. The image pyramid is used to calculate the reconstruction loss, where $ p $ is the pyramid index and $ P $ is number of pyramid layers.
We down-sample $ I_d $ and $ \hat{I}_d $ to 256 $\times$ 256, 128 $\times$ 128, 64 $\times$ 64 and 32 $\times$ 32 to build the image pyramid.

\subsubsection{Coefficient loss}

The coefficient loss is conducted to enforce the reenacted face to have the same identity as the source image and the same expression-and-pose as the driving image. Specifically, the identity, expression and pose coefficients of the generated image are extracted using the 3DMM regressor~\cite{zhu2017face}. Then we enforce the generated image to have the same identity coefficients as source image and the same expression-and-pose coefficients as the driving image: 
\begin{equation}
\begin{aligned}
L_{c} (I_s, I_d,\hat{I_d}) &= \sum_{k=0}^{50} || c_i(I_s)_k - c_i(\hat{I_d})_k || 
+ \sum_{k=0}^{51} || c_e(I_d)_k - c_e(\hat{I_d})_k || \\
&+ \sum_{k=0}^{12} || p(I_d)_k - p(\hat{I_d})_k ||
\end{aligned}
\end{equation}
where $ c_i(\cdot) $, $ c_e(\cdot) $ and $ p(\cdot) $ denote the functions that extract identity, expression and pose coefficients, respectively. $k$ denotes index of vector.

\subsubsection{Feature matching loss}

Following ~\cite{wang2018high}, we also add the feature matching loss to stabilize the training, so as to make the generator produce natural statistics at multiple feature layers:
\begin{equation}
\begin{aligned}
L_{FM} (I_d,\hat{I_d}) = \sum_{i=1}^T \frac{1}{N_i} [ || D^{i}(I_d) - D^i(\hat{I}_d) ||],
\end{aligned}
\end{equation}
where $T$ is the number of layers, $N_i$ denotes the
number of elements in each layer, if the layer output $D^i$ is a $64x64$ feature map, then $N_i$ is $64x64=4096$. $D^i$ denotes the feature map of $i$-th layer in the discriminator $D$.

\subsubsection{Adversarial Loss}

For the adversarial training, we adopt WGAN-GP~\cite{gulrajani2017improved}, whose adversarial loss is defined as:
\begin{equation}
\begin{aligned}
L_{adv}(I_d,\hat{I}_d) = \mathbb{E}_{\hat{I}_d \sim \mathbb{P}_{\hat{I}_d}} [D(\hat{I}_d)]  
- \mathbb{E}_{I_d \sim \mathbb{P}_{I_d}} [D(I_d)] \\
+ \lambda \mathbb{E}_{\hat{x} \sim \mathbb{P}_{\hat{x}}} [ || (\nabla_{\hat{x}}D(\hat{x})||_2 - 1)^2 ],
\end{aligned}
\end{equation}
where $ \hat{x} $ is uniformly sampled along the straight lines between the points sampled from the driving image distribution $ \mathbb{P}_{I_d} $ and the reenacted face images distribution $ \mathbb{P}_{\hat{I}_d} $.

\section{Implementation Details}

Before jointly training the motion net and reenacting module, we pre-train the 3DMM coefficients regressor~\cite{zhu2017face} using FaceWarehouse ~\cite{cao2013facewarehouse}, WFLW~\cite{wayne2018lab} and AFLW~\cite{koestinger2011annotated} datasets. The input image resolution is 256 $\times$ 256. For the 3DMM mesh, the vertex number is 53215, and the triangle face number is 105840. The Adam~\cite{kingma2014adam} is adopted as the optimizer.
We set the learning rate as $  2\times 10^{-4} $,
and the loss weights are 10,10,1,1 for $ \lambda_{rec} $, $ \lambda_{FM} $, $ \lambda_{c} $,and $ \lambda_{adv} $, respectively.

\section{Experiments}\label{Sec:5}

In this section, we evaluate the results of our proposed method and compare our results with state-of-the-art methods~\cite{xu2017face,zakharov2019fewshot,ha2019marionette,Siarohin_2019_NeurIPS}. For a fair comparison, all the methods are trained with the one-shot setting. Specifically, we adopt Openface 2.0~\cite{baltrusaitis2018openface} to estimate the facial landmarks for ~\cite{zakharov2019fewshot,ha2019marionette}. 
The public face datasets including Voxceleb1~\cite{nagrani2017voxceleb}, CelebV~\cite{wu2018reenactgan} and Faceforensics++~\cite{rossler2019faceforensics++} are employed to evaluate the results. We follow the evaluating protocol of ~\cite{ha2019marionette}.

\subsection{Qualitative Comparison}

\begin{figure*}[h]
	\centering
	\scalebox{.82}[.82]{
		\begin{tabular}{ccccccc}
			
			\includegraphics[scale=0.25]{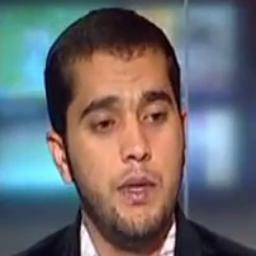}
			&\includegraphics[scale=0.25]{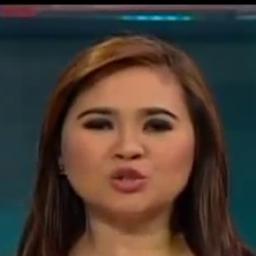}
			&\includegraphics[scale=0.25]{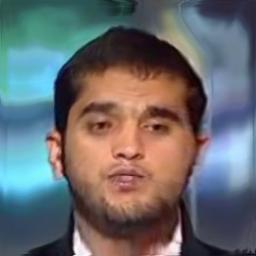}
			&\includegraphics[scale=0.25]{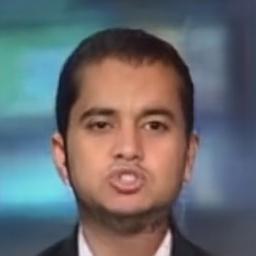}
			&\includegraphics[scale=0.25]{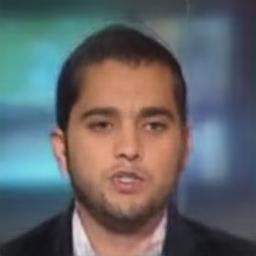}
			&\includegraphics[scale=0.25]{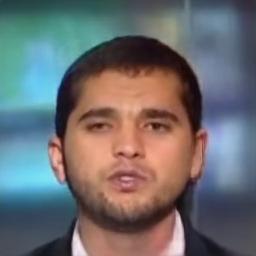}
			&\includegraphics[scale=0.25]{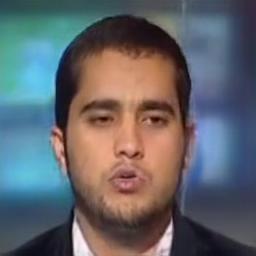}\\
			
			\includegraphics[scale=0.25]{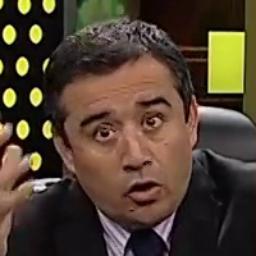}
			&\includegraphics[scale=0.25]{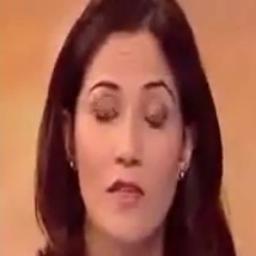}
			&\includegraphics[scale=0.25]{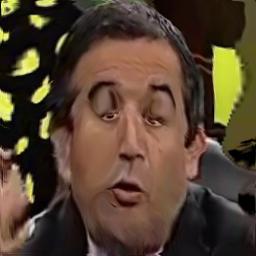}
			&\includegraphics[scale=0.25]{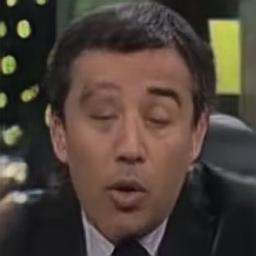}
			&\includegraphics[scale=0.25]{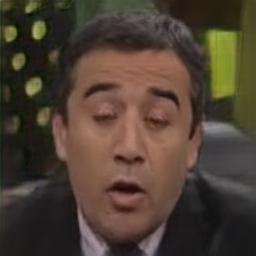}
			&\includegraphics[scale=0.25]{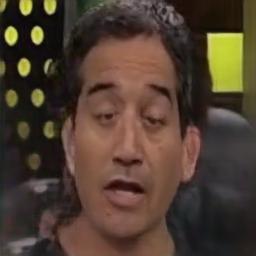}
			&\includegraphics[scale=0.25]{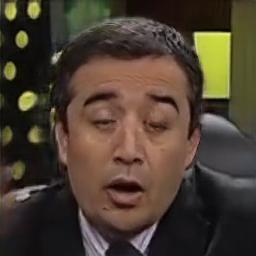}\\
			
			\includegraphics[scale=0.25]{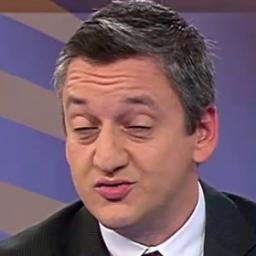}
			&\includegraphics[scale=0.25]{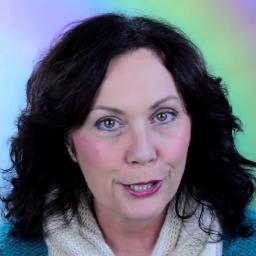}
			&\includegraphics[scale=0.25]{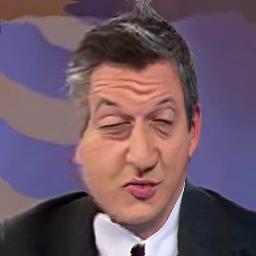}
			&\includegraphics[scale=0.25]{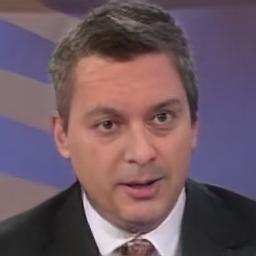}
			&\includegraphics[scale=0.25]{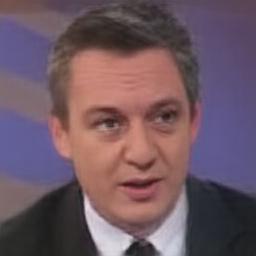}
			&\includegraphics[scale=0.25]{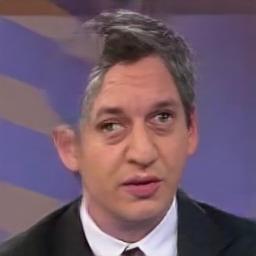}
			&\includegraphics[scale=0.25]{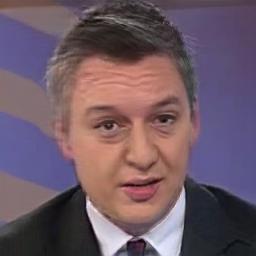}\\

			$I_s$ &$I_d$ &X2Face~\cite{wiles2018x2face} &NeuralHead-FF~\cite{zakharov2019fewshot}   &MarionNetTe~\cite{ha2019marionette} &FirstOrder~\cite{Siarohin_2019_NeurIPS} 
			&Ours\\
	\end{tabular}}
	\caption{Qualitative comparison with the state-of-the-art one-shot methods. The proposed method generates more natural-looking results with better preserved source identity and more accurate pose and expression.}
	\Description{comparison with the sota.}
	\label{fig:cmp}
\end{figure*}

Fig.~\ref{fig:cmp} shows the qualitative comparison of our method with recent methods~\cite{wiles2018x2face,zakharov2019fewshot,ha2019marionette,Siarohin_2019_NeurIPS}. The first column gives source images and the second column gives driving images. 
It is worth mentioning that the method of~\cite{zakharov2019fewshot} adopts meta-learning to generate faces, which requires fine-tuning for new-coming identities. We re-implement it using only the feed-forward network in the one-shot setting. The absolute motion transfer is performed for FirstOrder~\cite{Siarohin_2019_NeurIPS}.

As illustrated in Fig.~\ref{fig:cmp}, the source identities are not well preserved in the generated results of ~\cite{wiles2018x2face,zakharov2019fewshot,Siarohin_2019_NeurIPS}. 
The \textit{Landmark Transformer} mechanism proposed by MarionNET~\cite{ha2019marionette}
can alleviate this problem. Nevertheless, the result is still not satisfactory for the one-shot case. 
Our method can effectively preserve the source identities, as we explicitly exclude the driving face's identity information in the reconstructed driving face mesh.
Also, since we make use of the dense 3D meshes rather than sparse keypoints to learn the optical flow, our results can achieve more accurate poses and expressions. Furthermore, thanks to the full shape and pose information provided by dense meshes, our method can generate more natural-looking images with more details.

\begin{figure}[h]
	\centering
	\scalebox{.65}[.65]{
		\begin{tabular}{ccccccc}
			
			\includegraphics[scale=0.25]{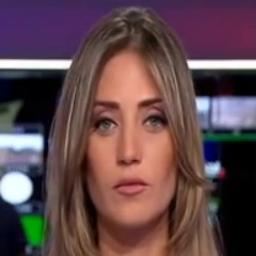}
			&\includegraphics[scale=0.25]{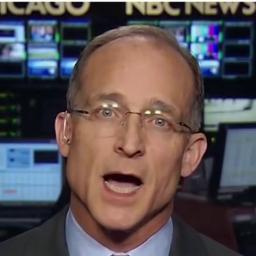}
			&\includegraphics[scale=0.25]{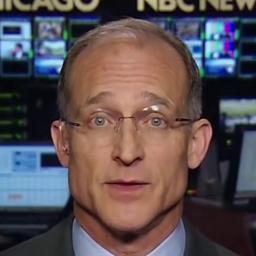}
			&\includegraphics[scale=0.25]{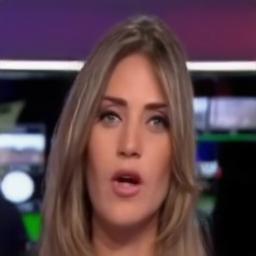}
			&\includegraphics[scale=0.25]{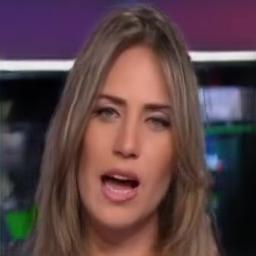}\\
			
			\includegraphics[scale=0.25]{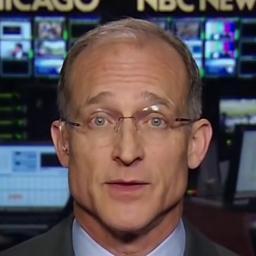}
			&\includegraphics[scale=0.25]{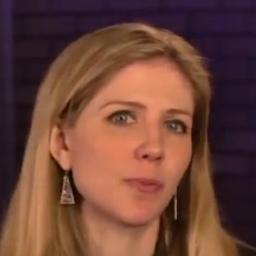}
			&\includegraphics[scale=0.25]{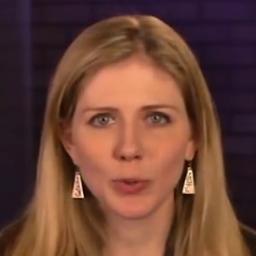}
			&\includegraphics[scale=0.25]{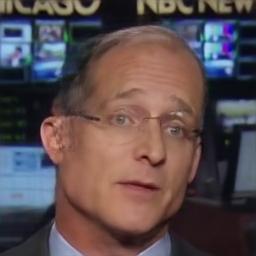}
			&\includegraphics[scale=0.25]{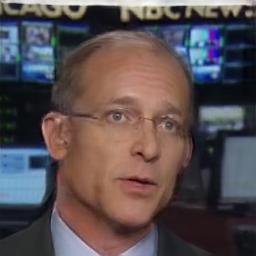}\\
			
			$I_s$ &$I_d$ &$ I_{init} $  &FirstOrder~\cite{Siarohin_2019_NeurIPS} 
			&Ours\\
	\end{tabular}}
	\setlength{\abovecaptionskip}{10pt}
	\caption{Qualitative comparison with FirstOrder~\cite{Siarohin_2019_NeurIPS} in the relative motion transfer setting. $ I_{init} $ provides the inital pose and expression for the driving face in ~\cite{Siarohin_2019_NeurIPS}. Images are from ~\cite{rossler2019faceforensics++}}
	\Description{comparison with sota.}
	\label{fig:relateive_cmp}
\end{figure}

Notice that FirstOrder~\cite{Siarohin_2019_NeurIPS} can achieve high-quality results if another driving image is provided as the initial expression and pose to perform relative motion transfer. To further validate the efficacy of our method, we compare our method (using one driving image) with the FirstOrder~\cite{Siarohin_2019_NeurIPS} (using two driving images). As shown in Fig.~\ref{fig:relateive_cmp}, we carefully select the initial driving image $ I_{init} $ so that $ I_{init}$ has the close pose and expression as $ I_s $.
We can see that our method can still yield more accurate results compared with ~\cite{Siarohin_2019_NeurIPS}. In the first row, the result of~\cite{Siarohin_2019_NeurIPS} has a smaller opening mouth, resulting in an expression that is not close to the expression of $ I_d $. In the second row, the head region is distorted in the result of~\cite{Siarohin_2019_NeurIPS}. In contrast, our results can well preserve the source identity and meanwhile owning a similar pose and expression as the driving image.

\subsection{Quantitative Comparison}

\begin{table}[h]
	\caption{Quantitative comparison in the self-reenactment setting. Up/down arrows correspond to higher/lower values for better performance. Bold and underlined numbers correspond to the best and the second-best values of each metric respectively.}
	\label{tab:Quantitative_Comparison}
	\scalebox{0.85}[0.85]{
		\begin{tabular}{cccccc}
			\toprule
			Model &CSIM$\uparrow$ &SSIM$\uparrow$ &PSNR$\uparrow$ &PRMSE$\downarrow$ &AUCON$\uparrow$\\
			\midrule
			
			\multicolumn{6}{c}{VoxCeleb1~\cite{nagrani2017voxceleb}}\\
			\midrule
			
			X2face\cite{wiles2018x2face} & 0.689 &0.719 &22.537 &3.26 &0.813\\
			
			NeuralHead-FF~\cite{zakharov2019fewshot} & 0.229 &0.635 &20.818 &3.76 &0.791\\
			
			MarioNETte~\cite{ha2019marionette} & 0.755 &\textbf{0.744} &23.244 &\textbf{3.13} &0.825\\
			
			FirstOrder~\cite{Siarohin_2019_NeurIPS} &\underline{0.813} &0.723 &\underline{30.182} &3.79 &\underline{0.886}\\
			
			Ours &\textbf{0.822} &\underline{0.739} &\textbf{30.394} &\underline{3.20} &\textbf{0.887} \\
			
			\midrule
			\multicolumn{6}{c}{Faceforensics++~\cite{rossler2019faceforensics++}}\\
			\midrule
			
			X2face~\cite{wiles2018x2face} &0.787 &0.654 &30.545 &6.10 &0.799 \\
			NeuralHead-FF~\cite{zakharov2019fewshot} &0.751  &0.680 &27.973 &6.09 &0.747 \\
			MarioNETte~\cite{ha2019marionette} &0.881 &0.694 &27.968 &3.72 &0.743 \\
			FirstOrder~\cite{Siarohin_2019_NeurIPS} &\underline{0.887} &\textbf{0.698} &\underline{30.620} &\underline{3.15} &\underline{0.839} \\
			Ours &\textbf{0.894} &\underline{0.695} &\textbf{30.648} &\textbf{2.71} &\textbf{0.858} \\
			\bottomrule
	\end{tabular}}
\end{table}

\begin{table}[h]
	\caption{Quantitative comparison of reenacting a different identity.}
	\label{tab:DifferenceID_Comparison}
	\begin{tabular}{cccc}
		\toprule
		Model &CSIM$\uparrow$ &PRMSE$\downarrow$ &AUCON$\uparrow$\\
		\midrule
		
		\multicolumn{4}{c}{CelebV~\cite{wu2018reenactgan}}\\
		\midrule
		
		X2face\cite{wiles2018x2face} &0.450 &3.62 &0.679\\
		
		NeuralHead-FF~\cite{zakharov2019fewshot} &0.108 &\textbf{3.30} &\textbf{0.722} \\
		
		MarioNETte~\cite{ha2019marionette} &\underline{0.520} &\underline{3.41} &\underline{0.710}\\
		
		FirstOrder~\cite{Siarohin_2019_NeurIPS} & 0.462 &3.90 &0.667 \\
		
		Ours &\textbf{0.635} &\underline{3.41} &0.709   \\
		
		\midrule
		\multicolumn{4}{c}{Faceforensics++~\cite{rossler2019faceforensics++}}\\
		\midrule
		
		X2face~\cite{wiles2018x2face} &0.604 &9.80 &0.697\\
		
		NeuralHead-FF~\cite{zakharov2019fewshot} &0.381 &6.82 &0.730 \\
		
		MarioNETte~\cite{ha2019marionette} &\underline{0.620} &7.68 &0.710\\
		
		FirstOrder~\cite{Siarohin_2019_NeurIPS} &0.614 &\underline{6.62} &\underline{0.734} \\
		
		Ours &\textbf{0.738}  &\textbf{6.24} &\textbf{0.737} \\
		
		\bottomrule
	\end{tabular}
\end{table}

Following the work of~\cite{ha2019marionette}, we employ the following metrics to quantitatively evaluate the reenacted faces of different methods. Peak signal-to-noise ratio (PSNR)~\cite{huynh2008scope} and structural similarity index (SSIM)~\cite{wang2004image} are utilized to measure the low-level similarity between the reenacted face and the ground-truth face, which are only computed in the self-reenactment scenario since the ground-truth is inaccessible when reenacting a different person.
Then we evaluate the identity preservation by calculating the cosine similarity (CSIM) of identity vectors between the source image and the generated image. The identity vectors are extracted by the pre-trained state-of-the-art face recognition networks~\cite{deng2019arcface}.
To inspect the model's capability of properly reenacting the pose and expression of driving image, we calculate PRMSE~\cite{ha2019marionette} and AUCON~\cite{ha2019marionette} between the generated image and the driving image to measure the reenacted pose and expression respectively.

Table~\ref{tab:Quantitative_Comparison} lists the evaluation results of different models in the self-reenactment setting, and Table~\ref{tab:DifferenceID_Comparison} reports the evaluation results of reenacting a different person. For FirstOder\cite{Siarohin_2019_NeurIPS}, we use the source image as the initial driving image to perform relative motion transfer in Table~\ref{tab:Quantitative_Comparison}, while the absolute motion transfer is performed in Table~\ref{tab:DifferenceID_Comparison} as the lack of initial images.
Notably, our method achieves the best scores in CSIM across all datasets, indicating that our method can better preserve the source identity than other methods. Besides, our method outperforms others in PRMSE and AUCON in most datasets, which demonstrates that our method can more faithfully reenact the pose and expression of driving face. Also, our method has the best scores in PSNR and the second-best scores in SSIM, which implies that our method can generate images closes to real images. Finally, it is worth noting that, when reenacting a different person, our method outperforms other methods in all metrics at the Faceforensics++~\cite{rossler2019faceforensics++} dataset, which is a very challenging dataset because it contains the most different identities.

\subsection{Ablation Study}

\begin{figure*}[h]
	\centering
	\scalebox{.8}[.8]{
		\begin{tabular}{ccccccc}
				\includegraphics[scale=0.25]{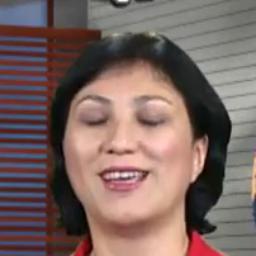}
				&\includegraphics[scale=0.25]{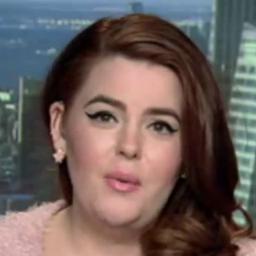}
				&\includegraphics[scale=0.25]{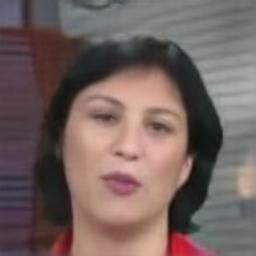}
				&\includegraphics[scale=0.25]{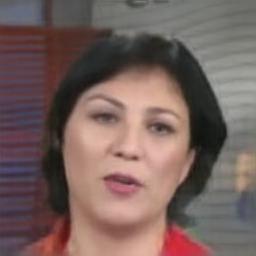}
				&\includegraphics[scale=0.25]{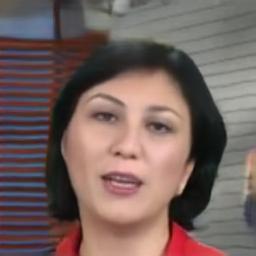}
				&\includegraphics[scale=0.25]{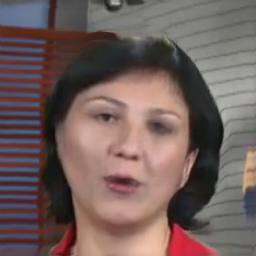}
				&\includegraphics[scale=0.25]{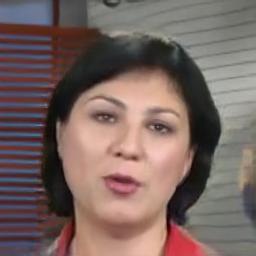}
				\\

				\includegraphics[scale=0.25]{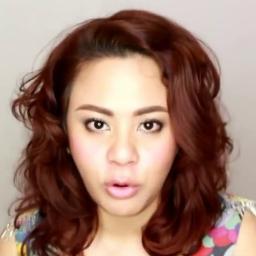}
				&\includegraphics[scale=0.25]{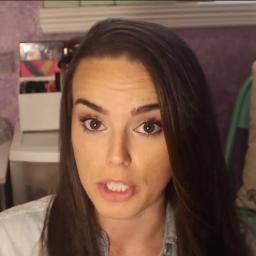}
				&\includegraphics[scale=0.25]{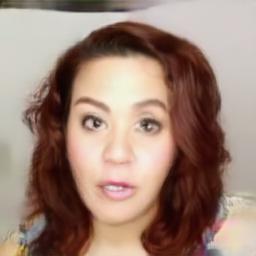}
				&\includegraphics[scale=0.25]{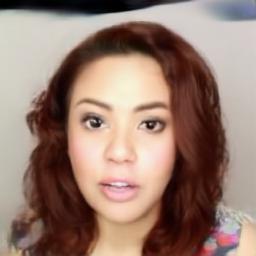}
				&\includegraphics[scale=0.25]{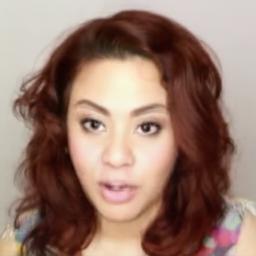}
				&\includegraphics[scale=0.25]{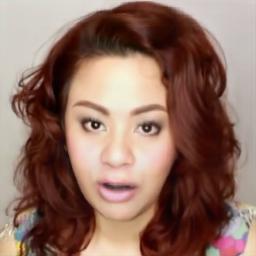}
				&\includegraphics[scale=0.25]{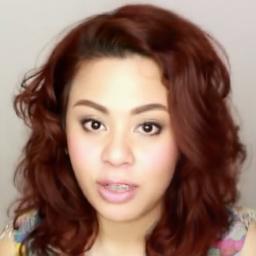}
				\\
				
				\includegraphics[scale=0.25]{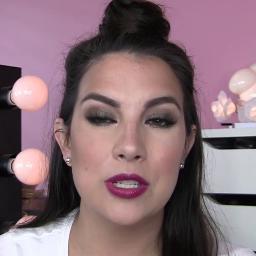}
				&\includegraphics[scale=0.25]{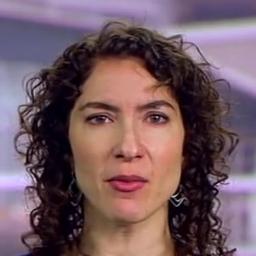}
				&\includegraphics[scale=0.25]{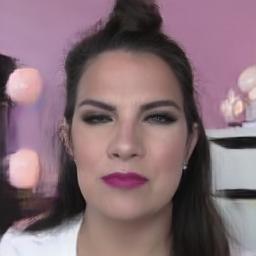}
				&\includegraphics[scale=0.25]{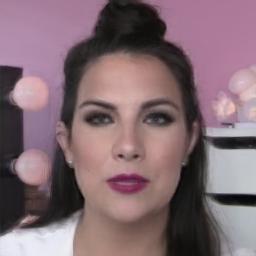}
				&\includegraphics[scale=0.25]{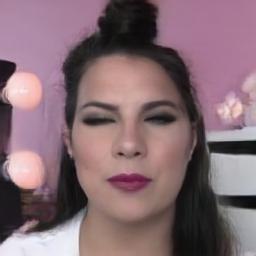}
				&\includegraphics[scale=0.25]{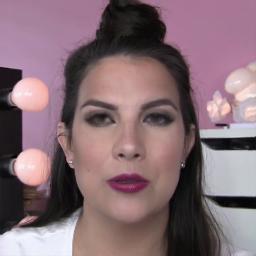}
				&\includegraphics[scale=0.25]{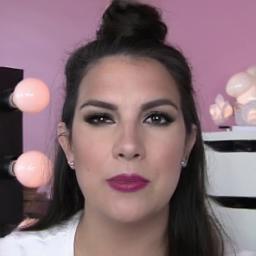}
				\\

			$I_s$ &$I_d$ &\textit{Baseline} &\textit{Baseline}+${L_{c}}$  &\textit{Baseline}+\textit{$ \mathcal{O} $}  &\textit{GCNs}-->\textit{CNNs}
			&Ours\\
			
	\end{tabular}}
	\vspace{-0.3cm}
	\caption{Qualitative ablation study on different examples. Our full model leads to better results than other variants.}
	\vspace{-0.3cm}
	\Description{Comparson with sota.}
	\label{fig:ablation}
\end{figure*}

\begin{table}[h]
	\caption{Quantitative ablation study for reenacting a different identity on the Faceforensics++ dataset~\cite{rossler2019faceforensics++}.}
	\label{tab:DifferenceID_Ablation}
	\begin{tabular}{cccccccccccc}
		\toprule
		\multicolumn{4}{c}{Model}        &CSIM$\uparrow$ &PRMSE$\downarrow$ &AUCON$\uparrow$ \\
		
		\midrule
		\multicolumn{4}{c}{\textit{Baseline}} &0.681 &7.453 &0.701  \\
		\multicolumn{4}{c}{\textit{Baseline}+${L_{c}}$} &0.687 &7.293 &0.715 \\
		\multicolumn{4}{c}{\textit{Baseline}+\textit{$ \mathcal{O} $}} &0.706 &7.030 &0.714\\
		\multicolumn{4}{c}{\textit{GCNs}-->\textit{CNNs}} &0.695  &7.51 &0.721\\
		\multicolumn{4}{c}{Ours} &\textbf{0.738}  &\textbf{6.24} &\textbf{0.737}\\
		\bottomrule
	\end{tabular}
	\vspace{-0.2cm}
\end{table}

We do the ablation study by evaluating the following variants of our method:
\begin{itemize}
	\item \textit{Baseline}. The simplest model trained without the occlusion net and ${L_{c}}$.
	\item \textit{Baseline}+${L_{c}}$. The coefficient loss is added to the \textit{Baseline}.
	\item \textit{Baseline}+\textit{$ \mathcal{O} $}. The occlusion net is added to the \textit{Baseline}.
	\item \textit{GCNs}-->\textit{CNNs}. To validate the effectiveness of learning the optical flow from 3D meshes using GCNs, we use the 2D rendered mesh images as input to learn the optical flow. Accordingly, the GCNs in the motion net are replaced with traditional CNNs. All the other components are the same as our model.
\end{itemize}

Fig.~\ref{fig:ablation} shows the qualitative results of the variants of our method. We can see that our full model presents the most realistic results. The coefficient loss can help to reduce the pose-and-expression error. The occlusion net can effectively improve image quality by reducing artifacts caused by occlusion (e.g. the head boundaries). Compared to learning the optical flow from rendered mesh images, our method that learns the optical flow from 3D meshes using GCNs can better preserve the source identity and obtain a closer pose and expression to the driving image.

We also report the quantitative results of the variants of our method on the ~\cite{rossler2019faceforensics++} dataset, as demonstrated in Table~\ref{tab:DifferenceID_Ablation}. Thanks to the explicit exclusion of driving identity in the reconstructed meshes, even the \textit{Baseline} surpasses the state-of-the-art methods in CSIM. 
With the help of the coefficient loss that penalizes the identity difference with the source and the expression-and-pose difference with the driving, the \textit{Baseline}+${L_{c}}$ obtains higher CSIM, AUCON, and lower PRMSE.
The \textit{Baseline}+${L_{c}}$ also achieves better results than the \textit{Baseline}, demonstrating the benefit of considering occlusions in face reenactment. The results of replacing GCNs with CNNs are much inferior compared to our full model using GCNs, revealing the importance of learning the optical flow from 3D meshes.

\subsection{User Study}

\begin{table}[h]
	\caption{User preferences of different methods on different datasets.}
	\label{tab:user_study}
	\scalebox{.9}[.9]{
	\begin{tabular}{cccccccccccc}
		\toprule
		Model  &VoxCeleb1~\cite{nagrani2017voxceleb} &Faceforensics++~\cite{rossler2019faceforensics++} \\
		\midrule
		
		X2face~\cite{wiles2018x2face}/Ours &0.016/0.984 &0.012/0.988 \\
		
		NeuralHead-FF~\cite{zakharov2019fewshot}/Ours &0.018/0.982 &0.036/0.964 \\
		
		MarioNETte~\cite{ha2019marionette}/Ours &0.126/0.874 &0.130/0.870\\
		
		FirstOrder~\cite{Siarohin_2019_NeurIPS}/Ours &0.124/0.876 &0.114/0.886 \\
		\bottomrule
	\end{tabular}}

\end{table}

We also conduct a user study to further assess the performance of our proposed model. 
We recruited 20 users (11 females) in the age range of 20-40. 
In every paired user study, each user needs to answer 50 questions. Each question consists of a source image, a driving image, the result of our method, and the result of one state-of-the-art method. The user needs to pick the result which she/he thinks has the better reenacted face. 
The user study results are reported in Table~\ref{tab:user_study}.
Table~\ref{tab:user_study} reports the selected probability ratio of each paired user study.
We can see that our method is clearly preferred over the state-of-art methods on different datasets.

\section{Application}

\begin{figure}[h]
	\centering
	\scalebox{.45}[.45]{
		\begin{tabular}{ccccccc}
			
			\includegraphics[scale=0.25]{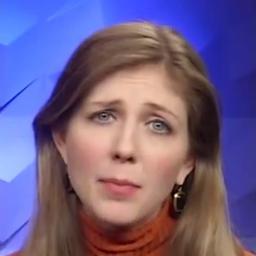}
			&\includegraphics[scale=0.25]{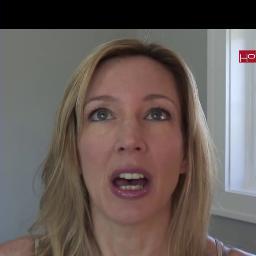}
			&\includegraphics[scale=0.25]{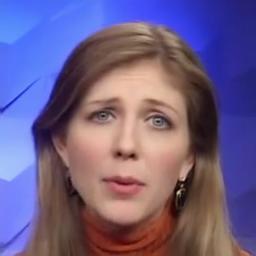}
			&\includegraphics[scale=0.25]{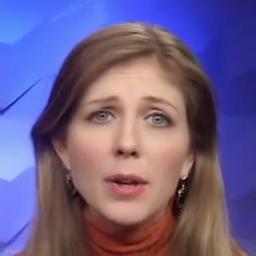}
			&\includegraphics[scale=0.25]{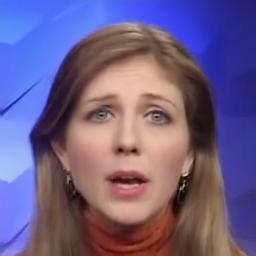}
			&\includegraphics[scale=0.25]{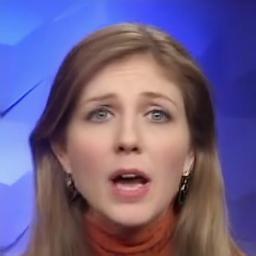}
			&\includegraphics[scale=0.25]{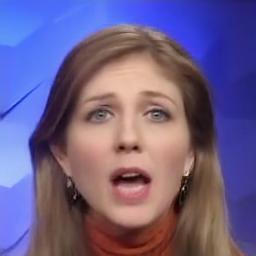}\\
			
			\includegraphics[scale=0.25]{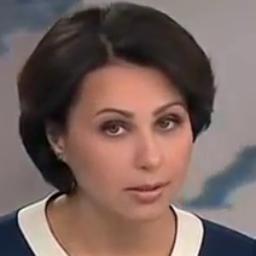}
			&\includegraphics[scale=0.25]{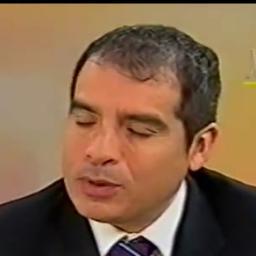}
			&\includegraphics[scale=0.25]{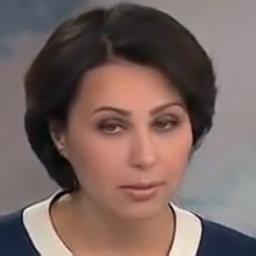}
			&\includegraphics[scale=0.25]{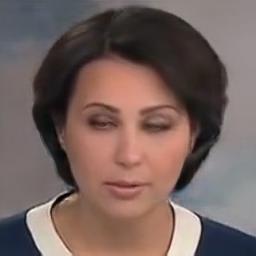}
			&\includegraphics[scale=0.25]{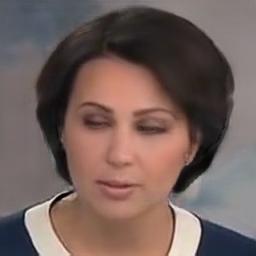}
			&\includegraphics[scale=0.25]{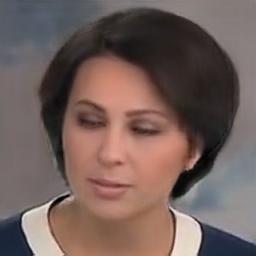}
			&\includegraphics[scale=0.25]{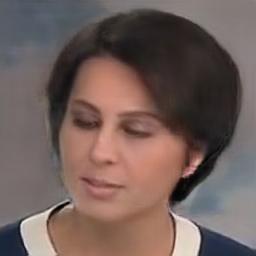}\\
			
			$I_s$ &$I_d$ &\textbf{ $\alpha$ = 0.2} &\textbf{ $\alpha$ = 0.4} &\textbf{ $\alpha$ = 0.6} &\textbf{ $\alpha$ = 0.8} &\textbf{ $\alpha$ = 1.0}\\
	\end{tabular}}
	\vspace{-0.3cm}
	\caption{Examples of expression and pose interpolation for intermediate face reenactment.}
	\vspace{-0.2cm}
	\label{fig:interp}
\end{figure}

\begin{figure}[h]
	\centering
	\scalebox{.63}[.63]{
		\begin{tabular}{ccccccc}
			
			\includegraphics[scale=0.25]{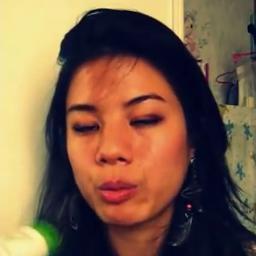}
			&\includegraphics[scale=0.25]{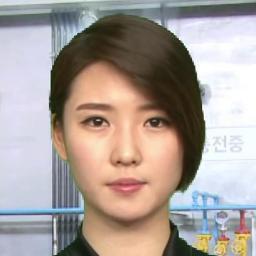}
			&\includegraphics[scale=0.25]{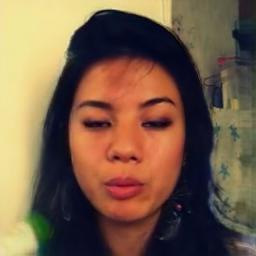}
			&\includegraphics[scale=0.25]{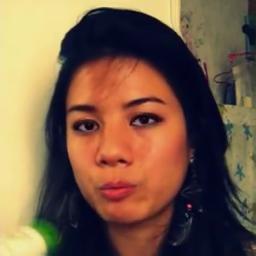}
			&\includegraphics[scale=0.25]{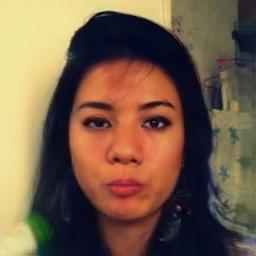}\\
			
			\includegraphics[scale=0.25]{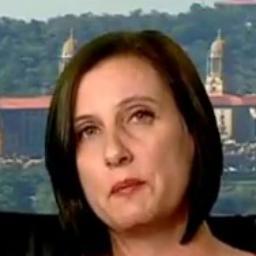}
			&\includegraphics[scale=0.25]{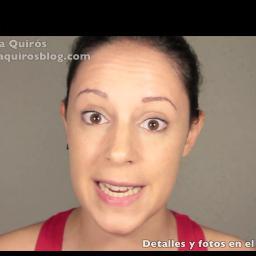}
			&\includegraphics[scale=0.25]{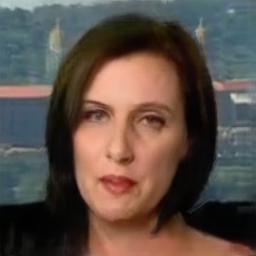}
			&\includegraphics[scale=0.25]{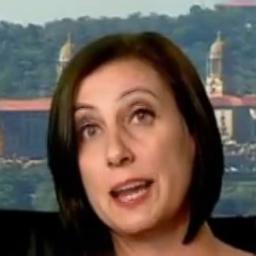}
			&\includegraphics[scale=0.25]{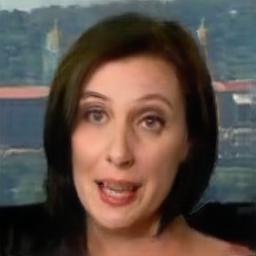}\\
			
			$I_s$ &$I_d$ &\textbf{only pose} &\textbf{only expression} &\textbf{pose + expression}\\
	\end{tabular}}
	\vspace{-0.3cm}
	\caption{Examples of disentangled reenactment of the pose or expression.}
	\vspace{-0.3cm}
	\label{fig:distangle}
\end{figure}

Our approach is guided by the reconstructed 3D meshes, in which the pose and expression parameters are extracted and controllable. Benefited by this, several interesting applications are enabled with our method. We show two applications here: 1) expression and pose interpolation for intermediate face reenactment and 2) disentangled reenactment of the pose or expression. In the expression and pose interpolation, we blend the source pose-and-expression parameters and the driving pose-and-expression parameters with the weight parameter $\alpha \in [0,1]$ to obtain
the intermediate driving pose-and-expression, which are used to build the driving mesh for a reenactment. As shown in Fig.~\ref{fig:interp}, when increasing $\alpha$ gradually from 0 to 1, the pose-and-expression of reenacted face are smoothly transferring from the source pose and expression to the driving pose and expression. 
This phenomenon also demonstrates our motion net learns a smooth optical flow for facial movement. In the application of disentangled reenactment of the pose or expression, we can see in Fig.~\ref{fig:distangle} that our approach can independently control the reenactment of face pose or expression, which is inaccessible for previous works~\cite{zakharov2019fewshot,wiles2018x2face,Siarohin_2019_CVPR,Siarohin_2019_NeurIPS}.

\section{Conclusions}
In this paper, we present a novel one-shot face reenactment framework, which animates a source image to another pose-and-expression.
Our method is guided by reconstructed meshes, which explicitly remove the driving identity information for better source identity preservation. Graph convolutional networks are deployed to learn the optical flow from dense meshes directly, which can obtain a more accurate pose and expression than learning from sparse keypoints.
To the best of our knowledge, we are the first to use graph convolutional networks to learn the facial movement from meshes for face reenactment from a single image. 
Compared with other methods, our method can generate more realistic and natural-looking results. 
In the future, we plan to explore the temporal consistency
in the network design to facilitate the face transfer in videos.

\begin{acks}
We thank anonymous reviewers for their valuable comments. This work is partially supported by the National Key Research \& Development Program of China (No. 2016YFB1001403), NSF China (No. 61772462, No. 61572429, No. U1736217) and the 100 Talents Program of Zhejiang University.
\end{acks}

\bibliographystyle{ACM-Reference-Format}
\bibliography{grimace}

\appendix
\newpage

\section{Architecture details}

\begin{figure*}[h]
	\centering
	\scalebox{1}[1]{
		\begin{tabular}{ccccccc}
			\includegraphics[scale=0.3]{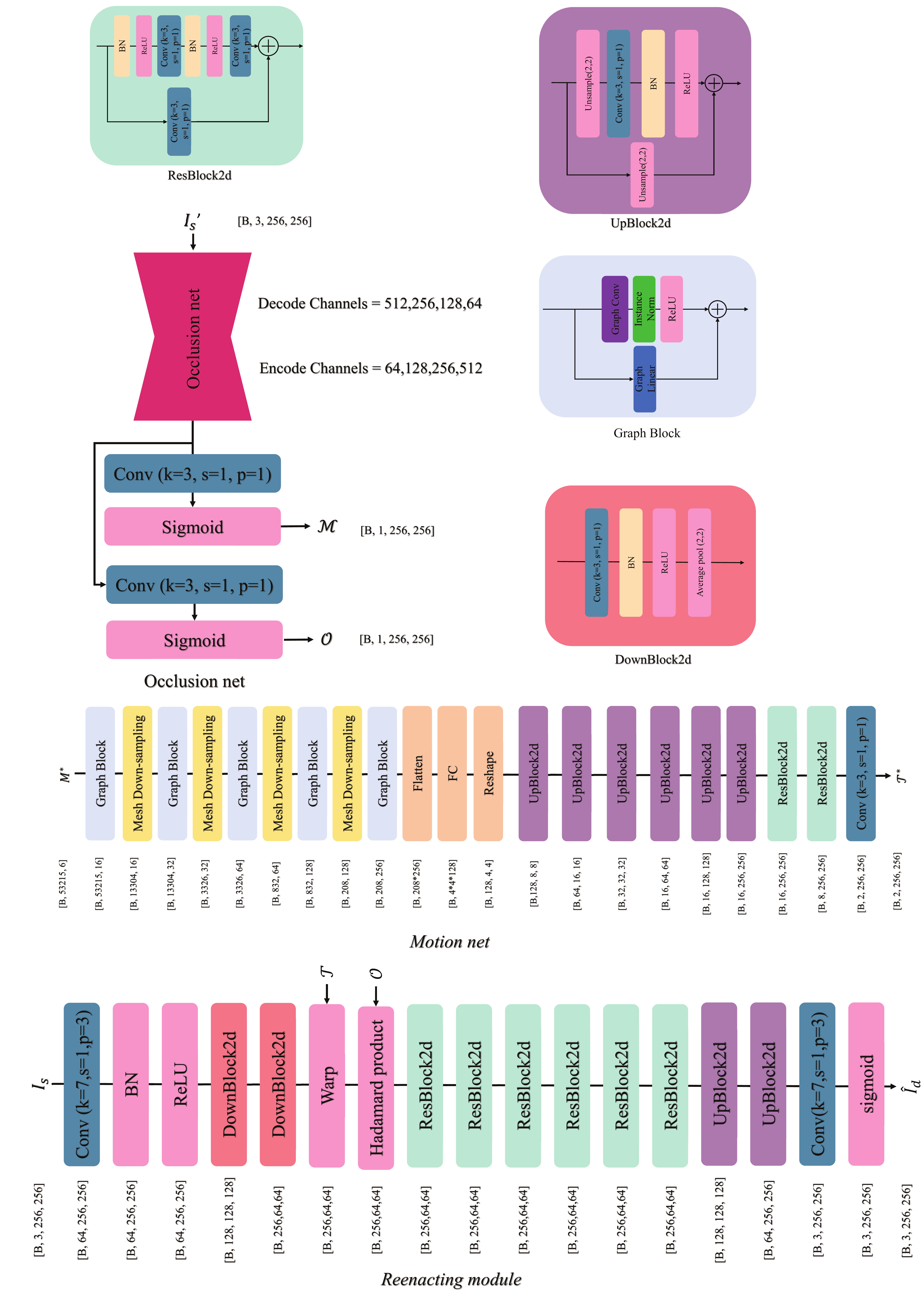}	
	\end{tabular}}
	\caption{Result of our method. The first row gives driving images and the first coloumn gives source images.}
	\label{fig:arch}
\end{figure*}

To make our work be an easily implemented work, the detailed architecture and output shape of each layer is shown as Fig.~\ref{fig:arch}.

\section{Inference Time}

In this section, we report the inference time for our approach when the resolution of the generated image is $256 \times 256$.
We test 100 times and report the average speed. We utilized Nvidia 2080Ti and Pytorch 1.0.1.post2. Table~\ref{tab:Inference_speed} lists the inference time break down of our methods.

\begin{table}[h]
	\caption{Inference speed of each component of our model.}
	\label{tab:Inference_speed}
	\begin{tabular}{cccccccccccc}
		\toprule
		Model  &Inference time (ms)\\
		\midrule
		
		Mesh regression &64 \\
		
		Motion net &37 \\
		
		Occlusion net &10 \\
		
		Encoder &7 \\
		
		Decoder &12 \\
		
		\bottomrule
	\end{tabular}
\end{table}

\section{Additional Examples}

\begin{figure*}[h]
	\centering
	\scalebox{1}[1]{
		\begin{tabular}{ccccccc}
			
			\includegraphics[scale=0.3]{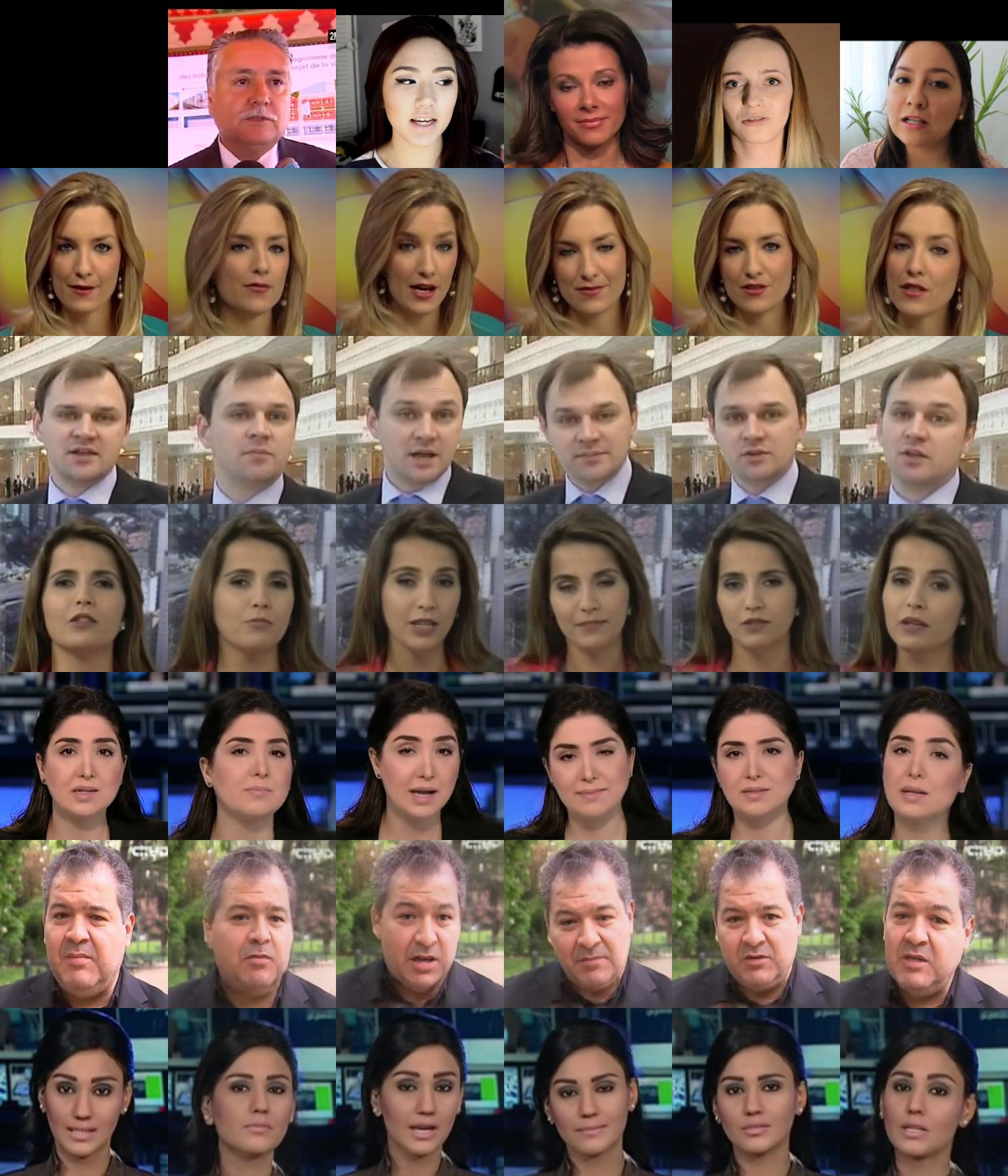}
			
	\end{tabular}}
	\caption{Result of our method. The first row gives driving images and the first coloumn gives source images.}
	\label{fig:multi}
\end{figure*}

We provide additional results to demonstrate the capability of our method. 
Qualitative comparison with the state-of-the-art one-shot methods shown as Fig.\ref{fig:cmp}, the proposed method generates more natural-looking results with better-preserved source identity and more accurate pose and expression
Fig.\ref{fig:ablation} demonstrates more ablation study result.
More results for interpolation and pose-and-expression disentanglement is shown at Fig.\ref{fig:interp} and Fig.\ref{fig:distangle} respectively.
Fig.\ref{fig:multi} demonstrates our method driving source image by some driving images of different identities, in which the first row gives driving images and the first column gives source images.

We believe a video is more convincing than words and images, and we thus additionally provide a video to evaluate the performance of our method. See the video \footnote{\url{youtu.be/Gv_pdnMfJMA}}.
In this video, the first row provides the driving frames, and the first column provides the source images. 
Our video results are relative smooth, however, as our method is designed for static images, some artifacts and temporal in-consistency problem happen in generated videos. we will fix this problem in our future work.

\section{Limitation}

\begin{figure*}[h]
	\centering
	\scalebox{0.4}[.4]{
		\includegraphics{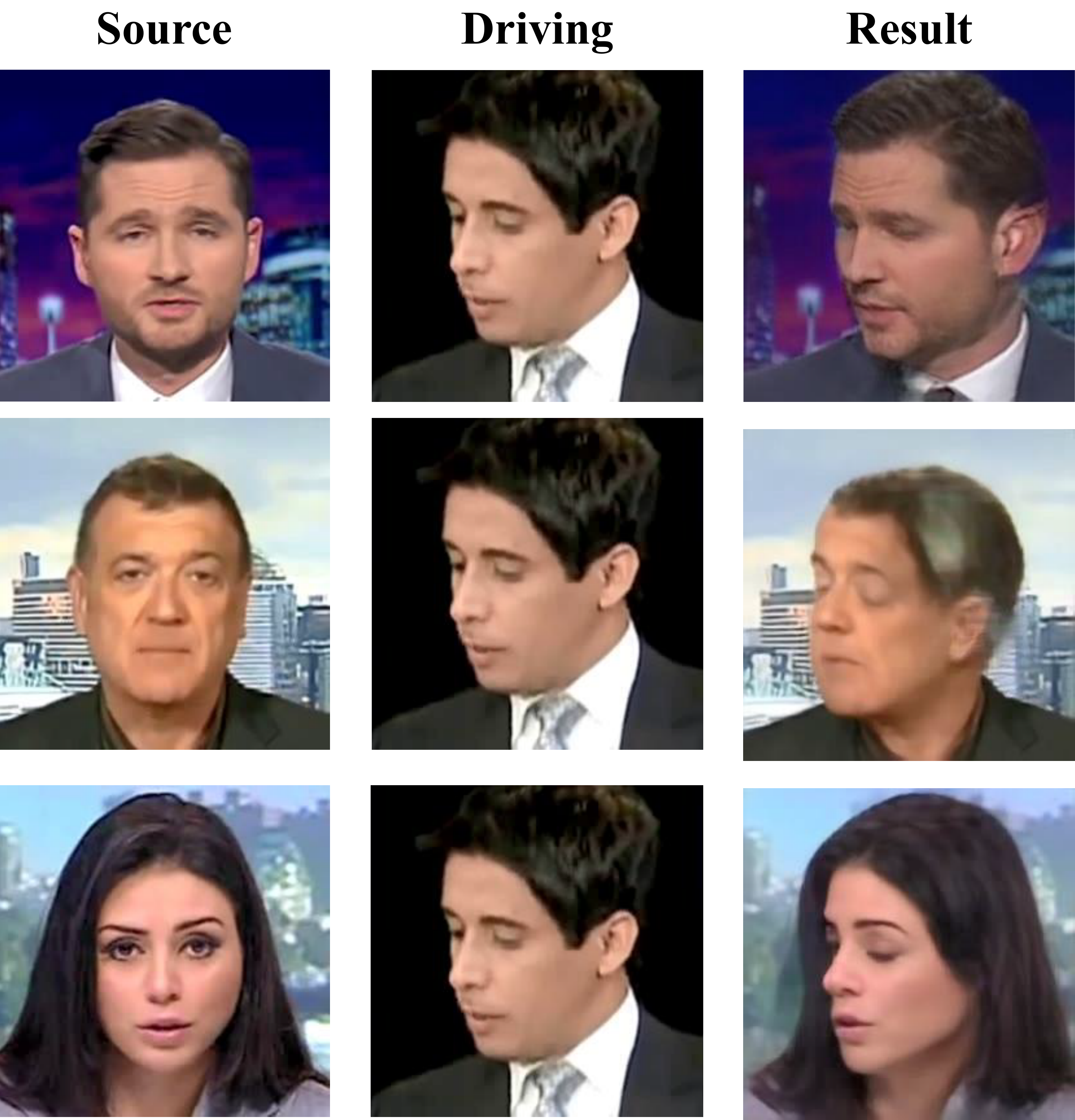}
	}
	\caption{Row 1-2 demonstrates failure cases generated by our method.
		Row 3 reveals that our method is able to work under extreme pose difference at sometime.}
	\label{fig:failure}
\end{figure*}

Fig.~\ref{fig:failure} row 1-2 reveals failure cases generated by our method, large pose difference between the source image, and the driving image is the main reason for the failures. 
However, row 3 of Fig.~\ref{fig:failure} also demonstrates that our framework is able to work under extreme pose difference at sometime.
Nevertheless, this limitation is also shared by state-of-art methods.

\begin{figure*}[hp]
	\centering
	\scalebox{.96}[.96]{
		\begin{tabular}{ccccccc}
			
			\includegraphics[scale=0.25]{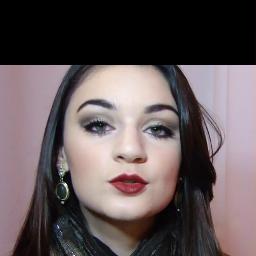}
			&\includegraphics[scale=0.25]{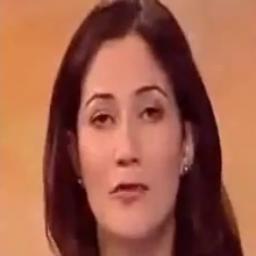}
			&\includegraphics[scale=0.25]{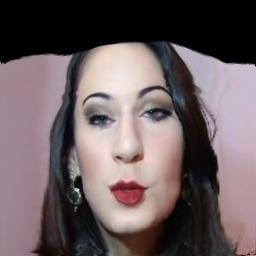}
			&\includegraphics[scale=0.25]{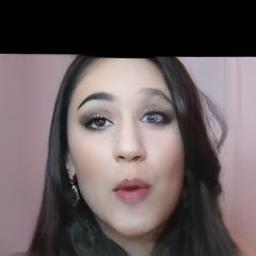}
			&\includegraphics[scale=0.25]{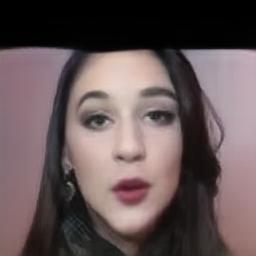}
			&\includegraphics[scale=0.25]{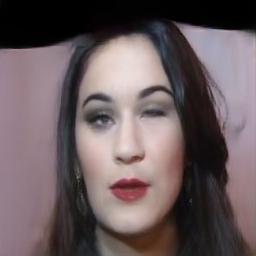}
			&\includegraphics[scale=0.25]{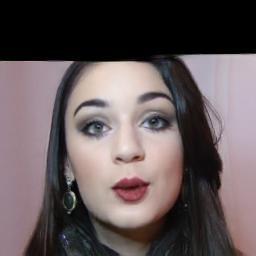}\\

			\includegraphics[scale=0.25]{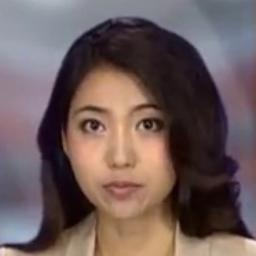}
			&\includegraphics[scale=0.25]{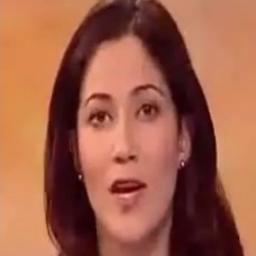}
			&\includegraphics[scale=0.25]{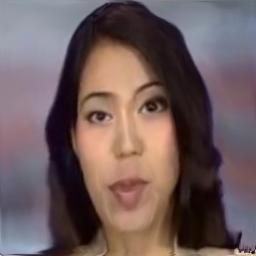}
			&\includegraphics[scale=0.25]{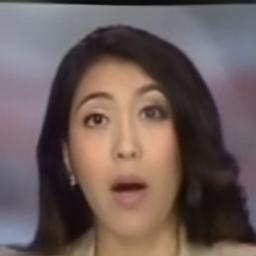}
			&\includegraphics[scale=0.25]{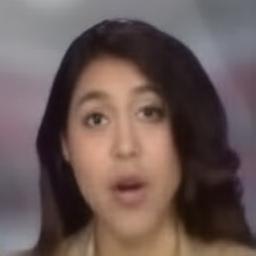}
			&\includegraphics[scale=0.25]{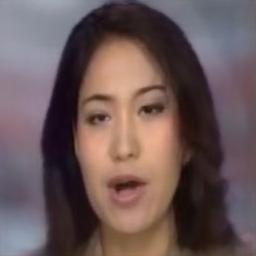}
			&\includegraphics[scale=0.25]{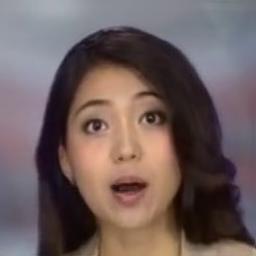}\\
			
			\includegraphics[scale=0.25]{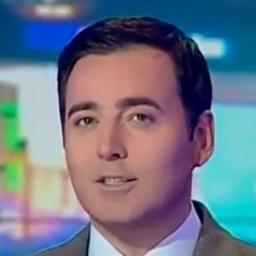}
			&\includegraphics[scale=0.25]{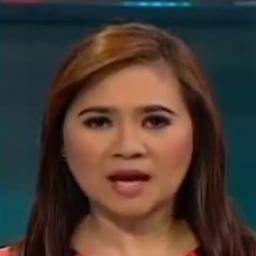}
			&\includegraphics[scale=0.25]{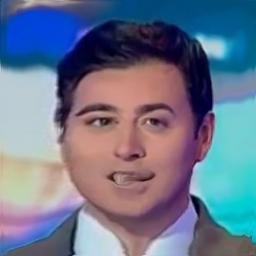}
			&\includegraphics[scale=0.25]{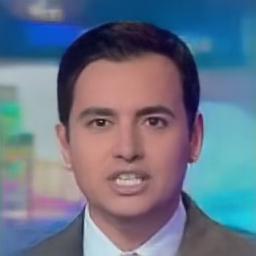}
			&\includegraphics[scale=0.25]{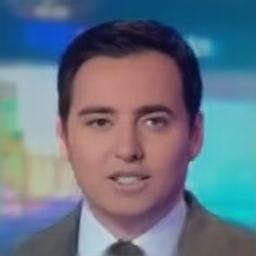}
			&\includegraphics[scale=0.25]{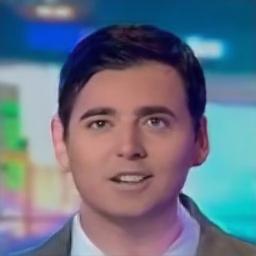}
			&\includegraphics[scale=0.25]{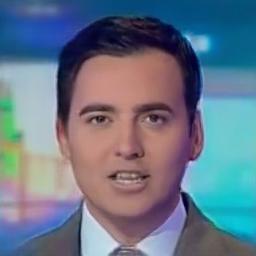}\\
			
			\includegraphics[scale=0.25]{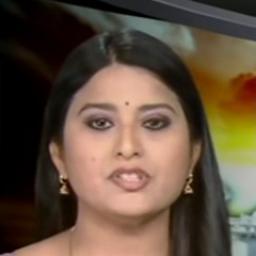}
			&\includegraphics[scale=0.25]{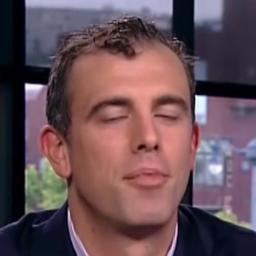}
			&\includegraphics[scale=0.25]{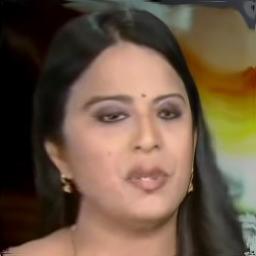}
			&\includegraphics[scale=0.25]{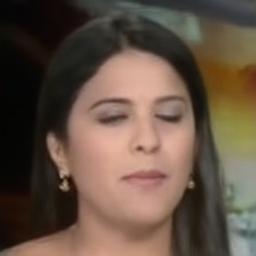}
			&\includegraphics[scale=0.25]{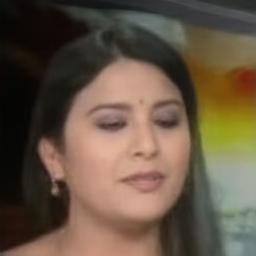}
			&\includegraphics[scale=0.25]{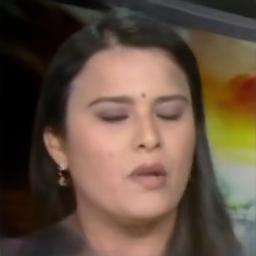}
			&\includegraphics[scale=0.25]{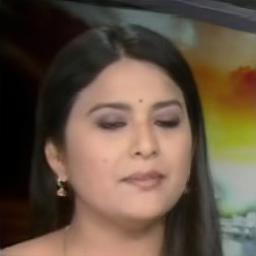}\\

			\includegraphics[scale=0.25]{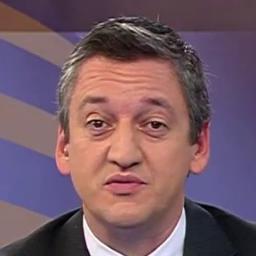}
			&\includegraphics[scale=0.25]{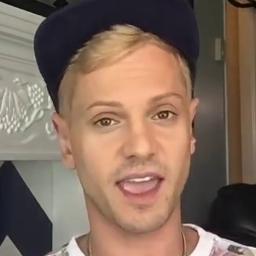}
			&\includegraphics[scale=0.25]{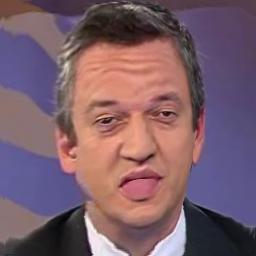}
			&\includegraphics[scale=0.25]{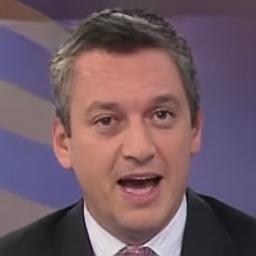}
			&\includegraphics[scale=0.25]{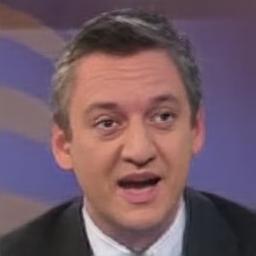}
			&\includegraphics[scale=0.25]{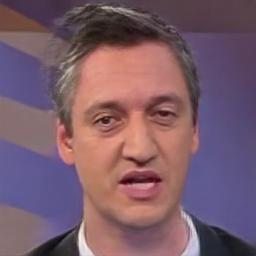}
			&\includegraphics[scale=0.25]{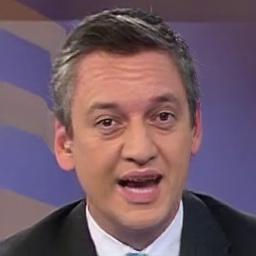}\\
			
			\includegraphics[scale=0.25]{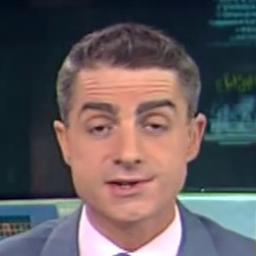}
			&\includegraphics[scale=0.25]{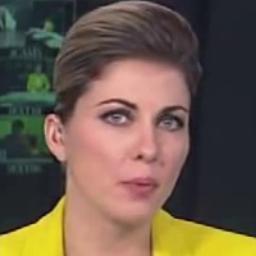}
			&\includegraphics[scale=0.25]{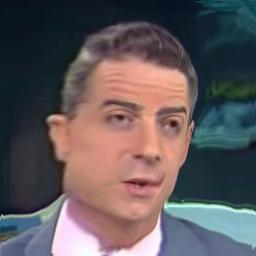}
			&\includegraphics[scale=0.25]{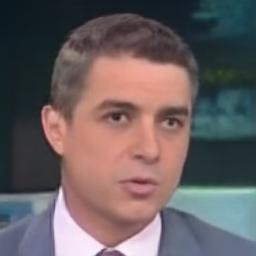}
			&\includegraphics[scale=0.25]{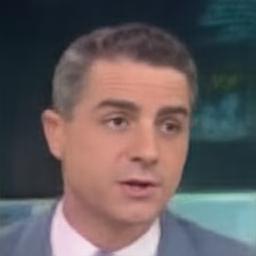}
			&\includegraphics[scale=0.25]{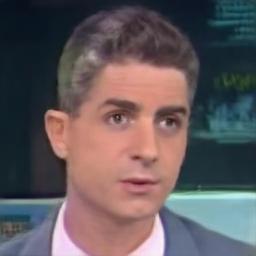}
			&\includegraphics[scale=0.25]{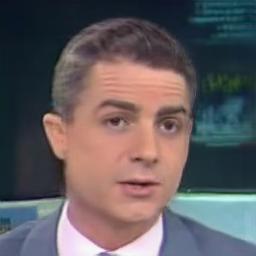}\\

			\includegraphics[scale=0.25]{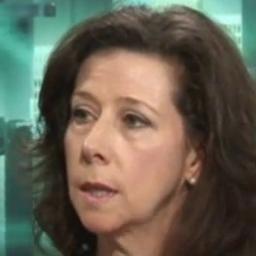}
			&\includegraphics[scale=0.25]{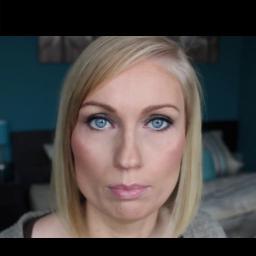}
			&\includegraphics[scale=0.25]{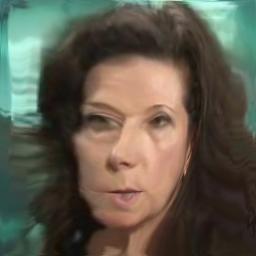}
			&\includegraphics[scale=0.25]{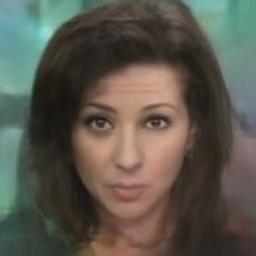}
			&\includegraphics[scale=0.25]{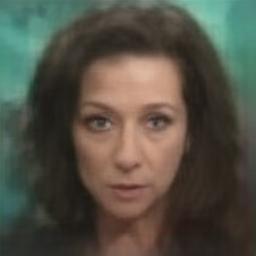}
			&\includegraphics[scale=0.25]{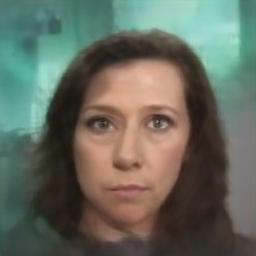}
			&\includegraphics[scale=0.25]{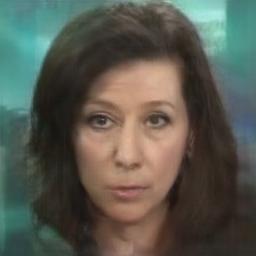}\\
			
			$I_s$ &$I_d$ &X2Face~\cite{wiles2018x2face} &NeuralHead-FF~\cite{zakharov2019fewshot}   &MarionNetTe~\cite{ha2019marionette} &FirstOrder~\cite{Siarohin_2019_NeurIPS} 
			&Ours\\
	\end{tabular}}
	\caption{Qualitative comparisons with state-of-the-art one-shot methods. The proposed method generates more natural-looking and sharp results compared to state of art methods.}
	\Description{comparison with the sota.}
	\label{fig:cmp}
\end{figure*}

\begin{figure*}[h]
	\centering
	\scalebox{.93}[.93]{
		\begin{tabular}{ccccccc}

			\includegraphics[scale=0.25]{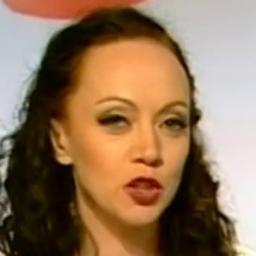}
			&\includegraphics[scale=0.25]{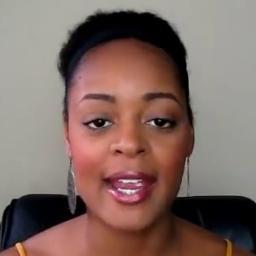}
			&\includegraphics[scale=0.25]{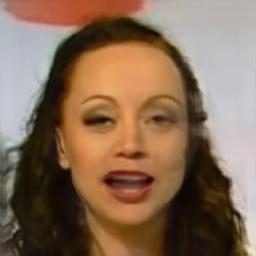}
			&\includegraphics[scale=0.25]{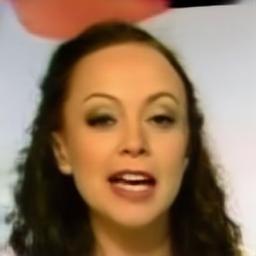}
			&\includegraphics[scale=0.25]{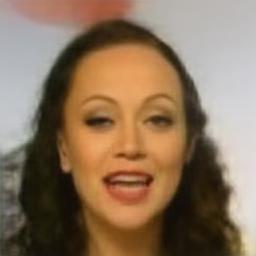}
			&\includegraphics[scale=0.25]{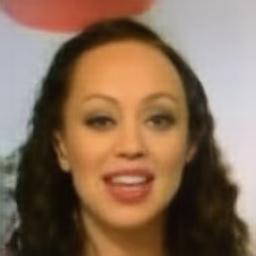}
			&\includegraphics[scale=0.25]{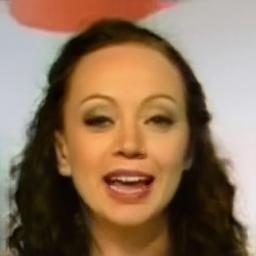}
			\\
			
			\includegraphics[scale=0.25]{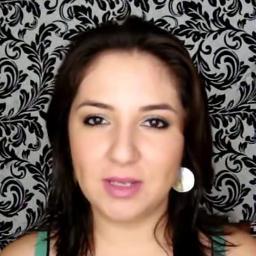}
			&\includegraphics[scale=0.25]{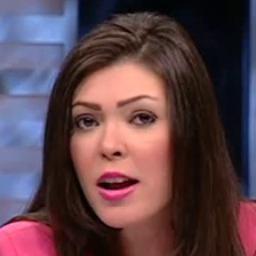}
			&\includegraphics[scale=0.25]{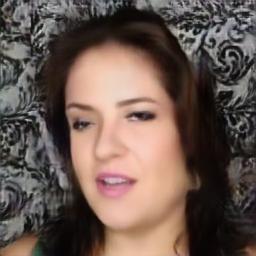}
			&\includegraphics[scale=0.25]{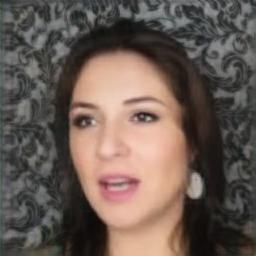}
			&\includegraphics[scale=0.25]{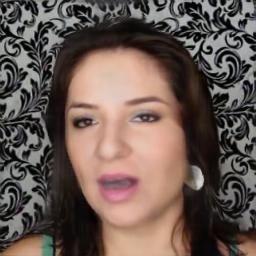}
			&\includegraphics[scale=0.25]{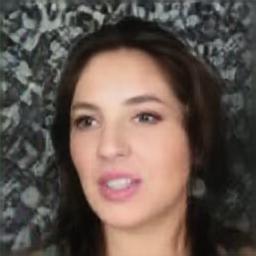}
			&\includegraphics[scale=0.25]{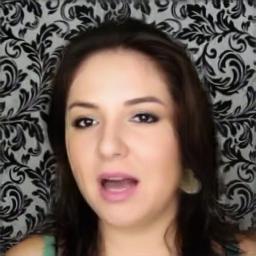}
			\\

			\includegraphics[scale=0.25]{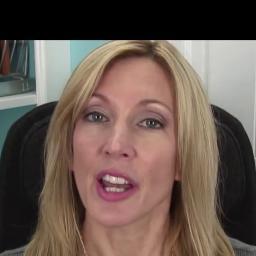}
			&\includegraphics[scale=0.25]{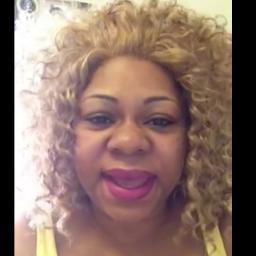}
			&\includegraphics[scale=0.25]{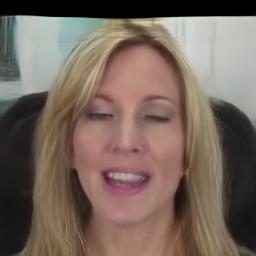}
			&\includegraphics[scale=0.25]{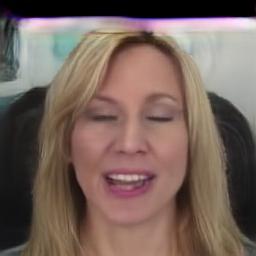}
			&\includegraphics[scale=0.25]{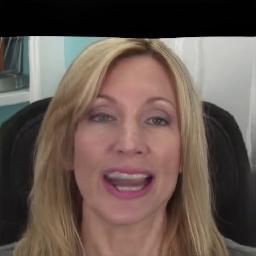}
			&\includegraphics[scale=0.25]{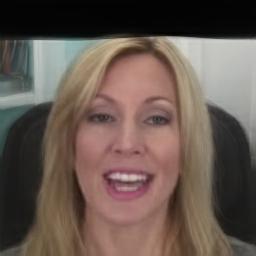}
			&\includegraphics[scale=0.25]{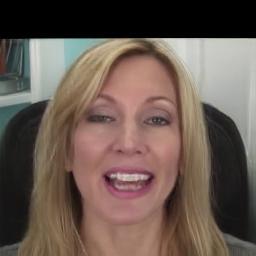}
			\\
			
			\includegraphics[scale=0.25]{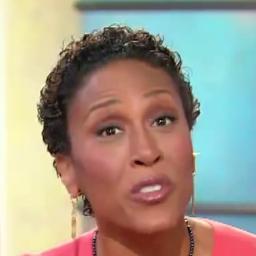}
			&\includegraphics[scale=0.25]{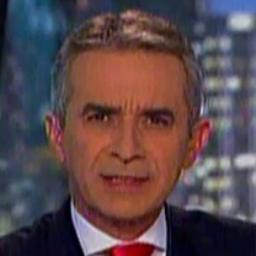}
			&\includegraphics[scale=0.25]{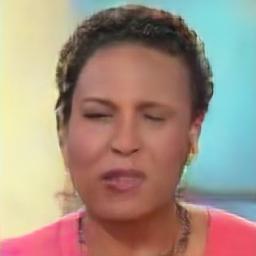}
			&\includegraphics[scale=0.25]{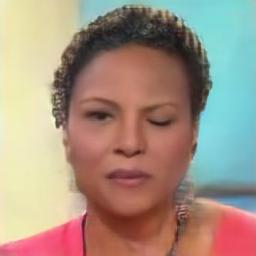}
			&\includegraphics[scale=0.25]{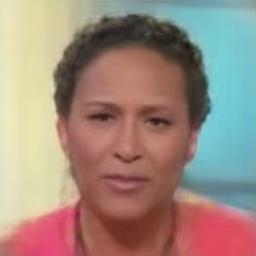}
			&\includegraphics[scale=0.25]{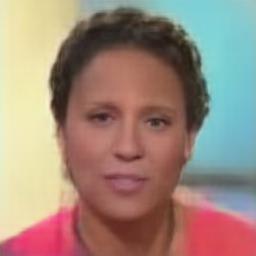}
			&\includegraphics[scale=0.25]{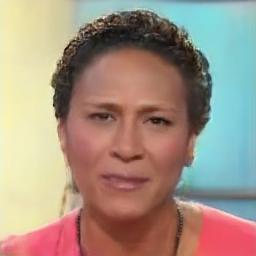}
			\\

			\includegraphics[scale=0.25]{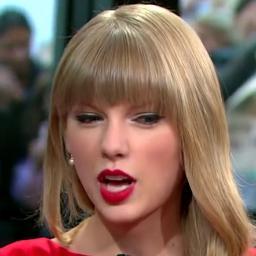}
			&\includegraphics[scale=0.25]{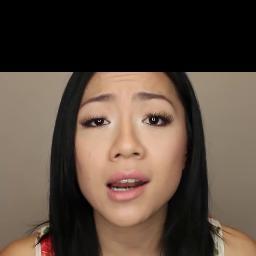}
			&\includegraphics[scale=0.25]{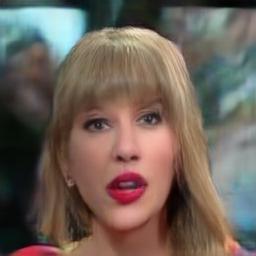}
			&\includegraphics[scale=0.25]{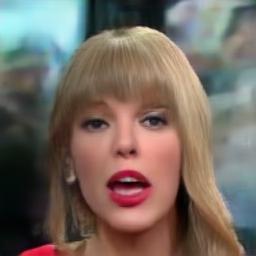}
			&\includegraphics[scale=0.25]{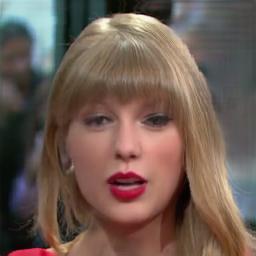}
			&\includegraphics[scale=0.25]{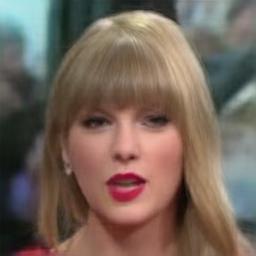}
			&\includegraphics[scale=0.25]{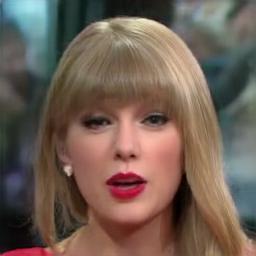}
			\\
			
			\includegraphics[scale=0.25]{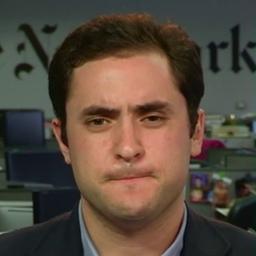}
			&\includegraphics[scale=0.25]{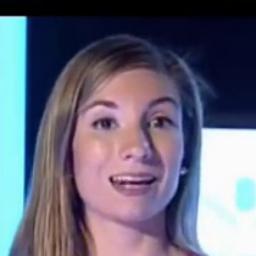}
			&\includegraphics[scale=0.25]{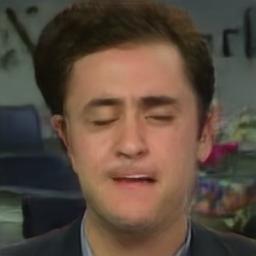}
			&\includegraphics[scale=0.25]{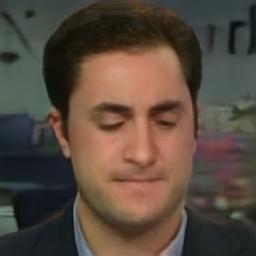}
			&\includegraphics[scale=0.25]{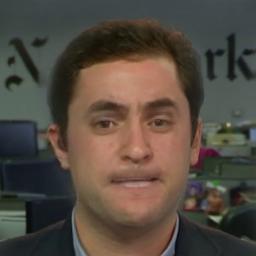}
			&\includegraphics[scale=0.25]{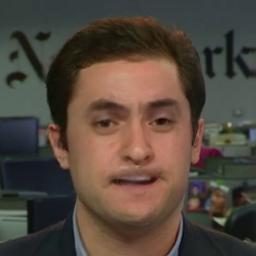}
			&\includegraphics[scale=0.25]{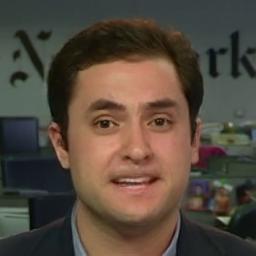}
			\\			
			$I_s$ &$I_d$ &\textit{Baseline} &\textit{Baseline}+${L_{c}}$  &\textit{Baseline}+\textit{$ \mathcal{O} $}  &\textit{GCNs}-->\textit{CNNs}
			&Ours\\
			
	\end{tabular}}
	\caption{Ablation study on different identities. Our model leads to a better result than others.}
	\Description{Comparson with sota.}
	\label{fig:ablation}
\end{figure*}

\begin{figure*}[h]
	\centering
	\scalebox{.93}[.93]{
		\begin{tabular}{ccccccc}
			
			\includegraphics[scale=0.25]{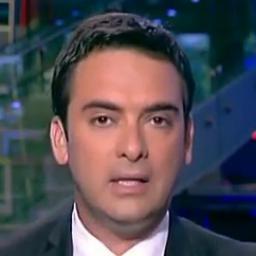}
			&\includegraphics[scale=0.25]{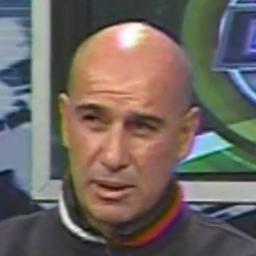}
			&\includegraphics[scale=0.25]{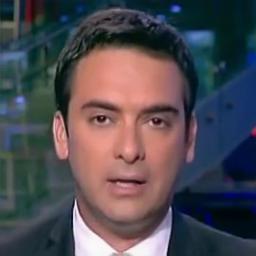}
			&\includegraphics[scale=0.25]{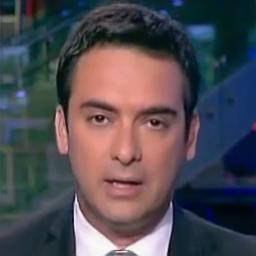}
			&\includegraphics[scale=0.25]{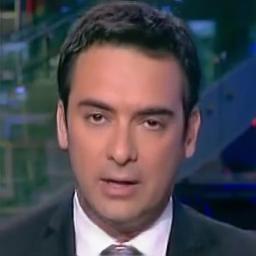}
			&\includegraphics[scale=0.25]{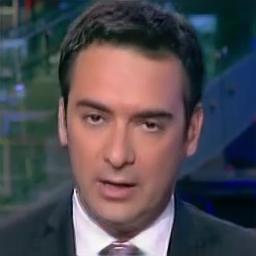}
			&\includegraphics[scale=0.25]{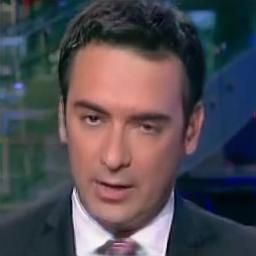}\\
			
			\includegraphics[scale=0.25]{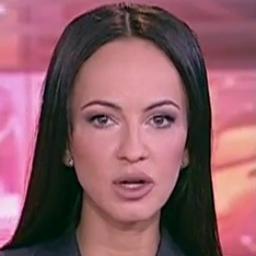}
			&\includegraphics[scale=0.25]{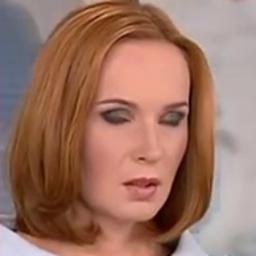}
			&\includegraphics[scale=0.25]{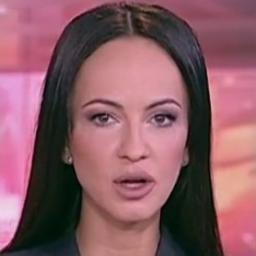}
			&\includegraphics[scale=0.25]{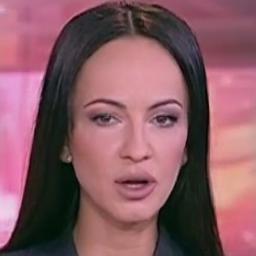}
			&\includegraphics[scale=0.25]{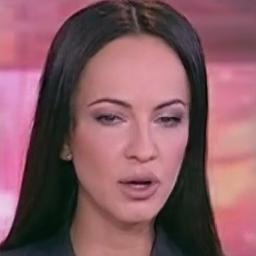}
			&\includegraphics[scale=0.25]{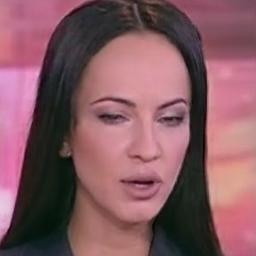}
			&\includegraphics[scale=0.25]{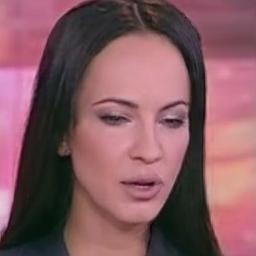}\\
			
			\includegraphics[scale=0.25]{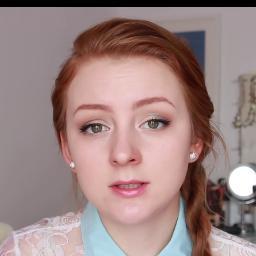}
			&\includegraphics[scale=0.25]{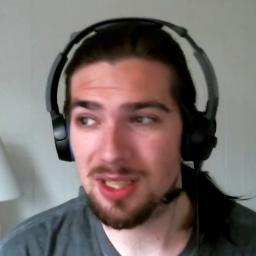}
			&\includegraphics[scale=0.25]{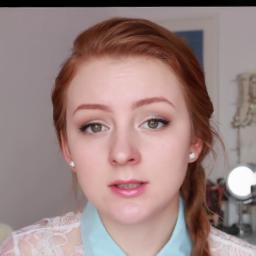}
			&\includegraphics[scale=0.25]{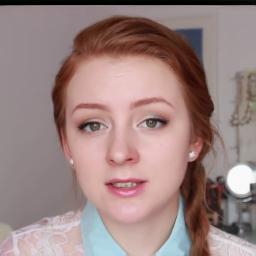}
			&\includegraphics[scale=0.25]{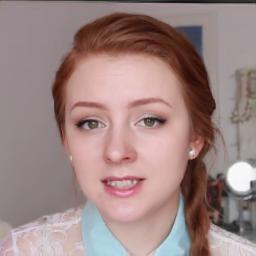}
			&\includegraphics[scale=0.25]{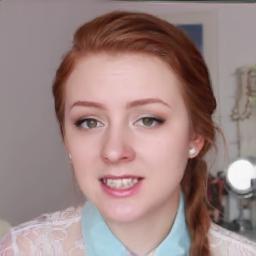}
			&\includegraphics[scale=0.25]{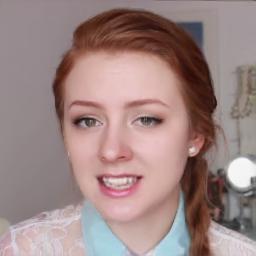}\\
			
			\includegraphics[scale=0.25]{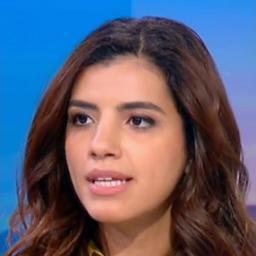}
			&\includegraphics[scale=0.25]{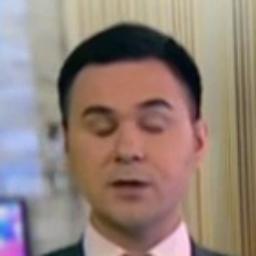}
			&\includegraphics[scale=0.25]{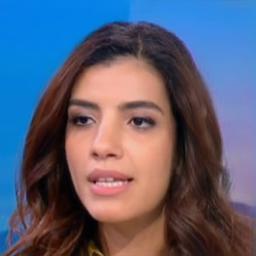}
			&\includegraphics[scale=0.25]{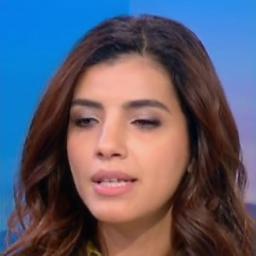}
			&\includegraphics[scale=0.25]{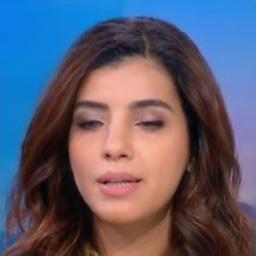}
			&\includegraphics[scale=0.25]{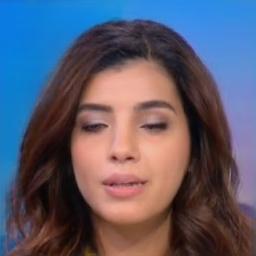}
			&\includegraphics[scale=0.25]{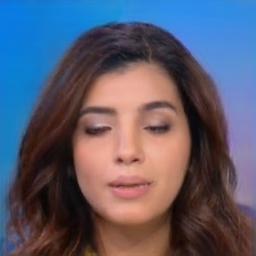}\\
			
			\includegraphics[scale=0.25]{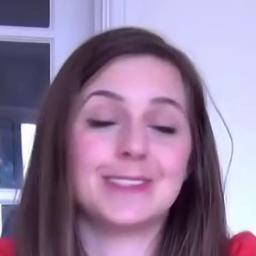}
			&\includegraphics[scale=0.25]{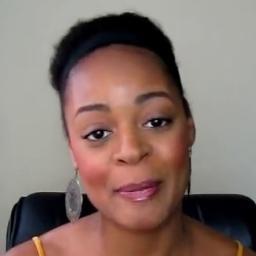}
			&\includegraphics[scale=0.25]{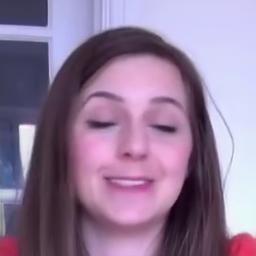}
			&\includegraphics[scale=0.25]{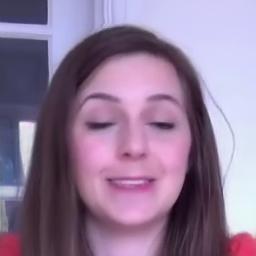}
			&\includegraphics[scale=0.25]{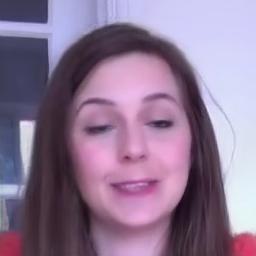}
			&\includegraphics[scale=0.25]{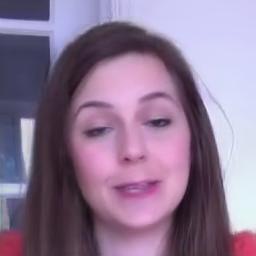}
			&\includegraphics[scale=0.25]{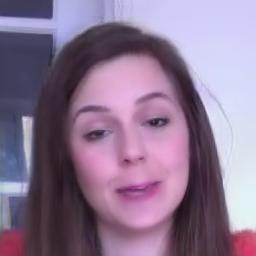}\\
			
			\includegraphics[scale=0.25]{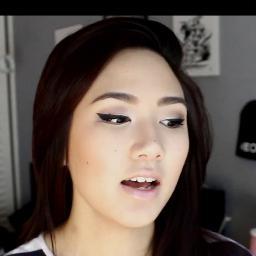}
			&\includegraphics[scale=0.25]{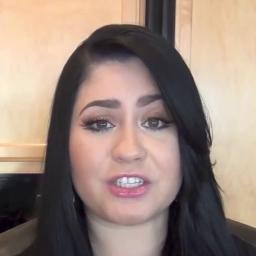}
			&\includegraphics[scale=0.25]{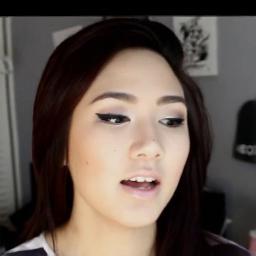}
			&\includegraphics[scale=0.25]{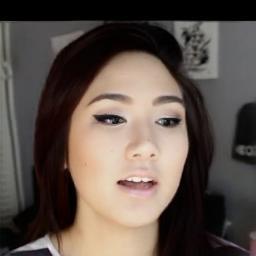}
			&\includegraphics[scale=0.25]{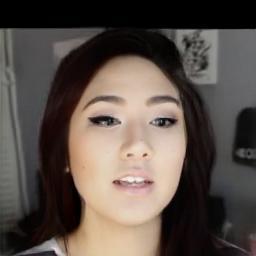}
			&\includegraphics[scale=0.25]{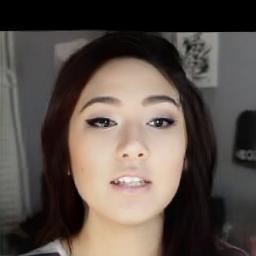}
			&\includegraphics[scale=0.25]{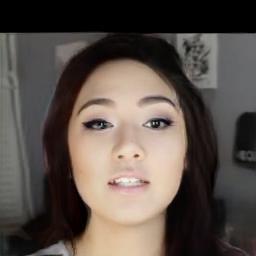}\\

			\includegraphics[scale=0.25]{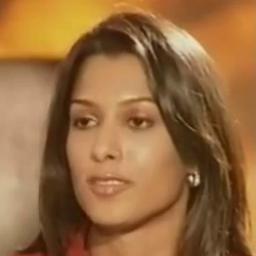}
			&\includegraphics[scale=0.25]{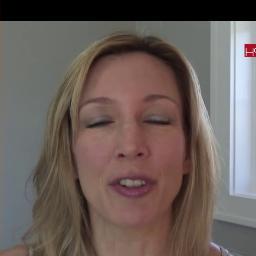}
			&\includegraphics[scale=0.25]{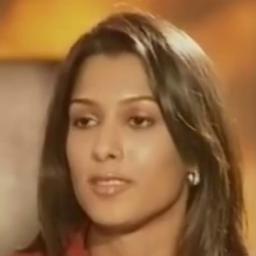}
			&\includegraphics[scale=0.25]{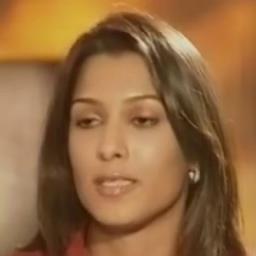}
			&\includegraphics[scale=0.25]{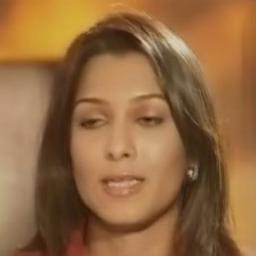}
			&\includegraphics[scale=0.25]{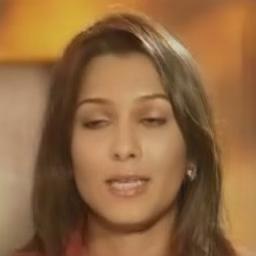}
			&\includegraphics[scale=0.25]{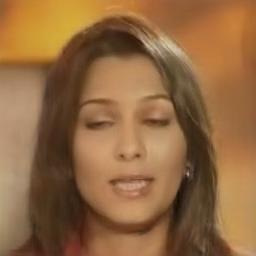}\\
			
			\includegraphics[scale=0.25]{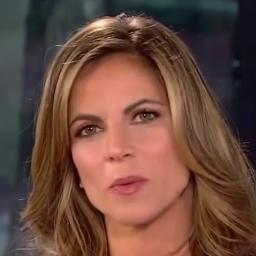}
			&\includegraphics[scale=0.25]{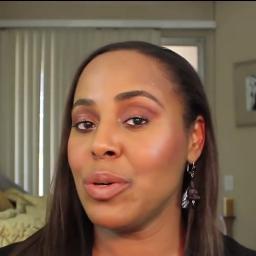}
			&\includegraphics[scale=0.25]{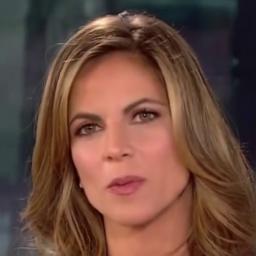}
			&\includegraphics[scale=0.25]{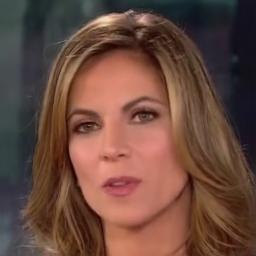}
			&\includegraphics[scale=0.25]{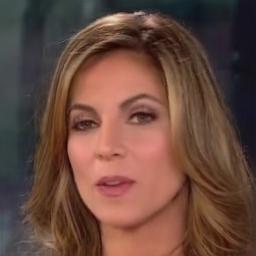}
			&\includegraphics[scale=0.25]{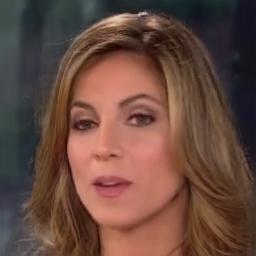}
			&\includegraphics[scale=0.25]{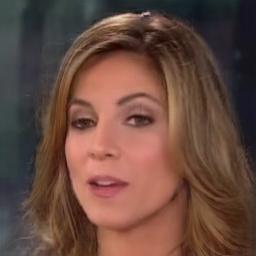}\\
			
			\includegraphics[scale=0.25]{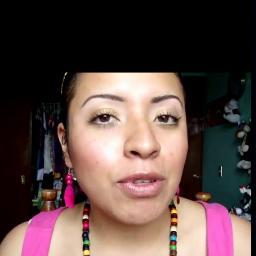}
			&\includegraphics[scale=0.25]{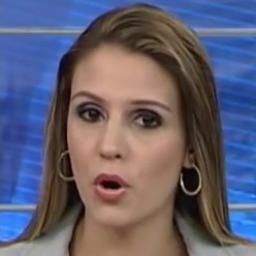}
			&\includegraphics[scale=0.25]{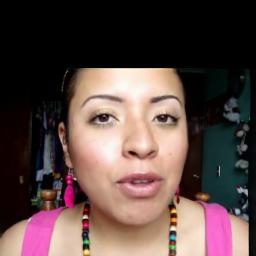}
			&\includegraphics[scale=0.25]{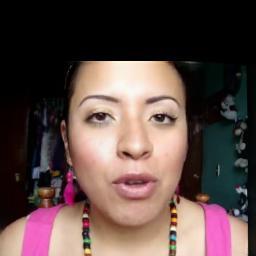}
			&\includegraphics[scale=0.25]{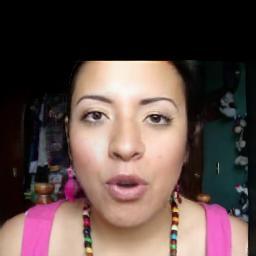}
			&\includegraphics[scale=0.25]{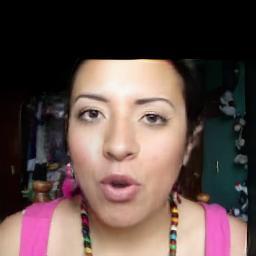}
			&\includegraphics[scale=0.25]{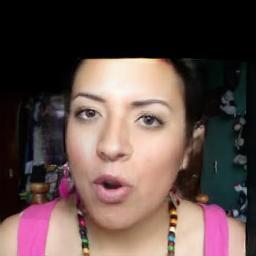}\\			
			$I_s$ &$I_d$ &\textbf{ $\alpha$ = 0.2} &\textbf{ $\alpha$ = 0.4} &\textbf{ $\alpha$ = 0.6} &\textbf{ $\alpha$ = 0.8} &\textbf{ $\alpha$ = 1.0}\\
	\end{tabular}}
	\caption{Interpolation of the proposed approach.}
	\label{fig:interp}
\end{figure*}

\begin{figure*}[h]
	\centering
	\scalebox{1.3}[1.3]{
		\begin{tabular}{ccccccc}
			
			\includegraphics[scale=0.25]{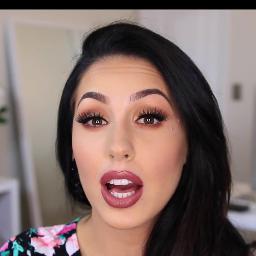}
			&\includegraphics[scale=0.25]{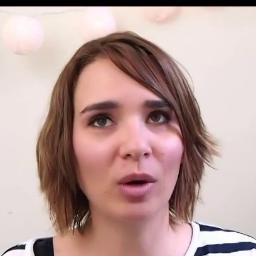}
			&\includegraphics[scale=0.25]{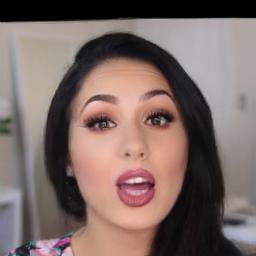}
			&\includegraphics[scale=0.25]{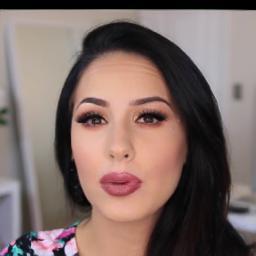}
			&\includegraphics[scale=0.25]{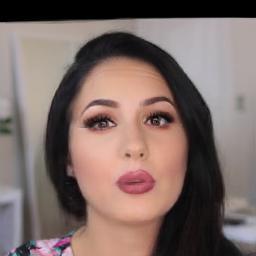}\\
			
			\includegraphics[scale=0.25]{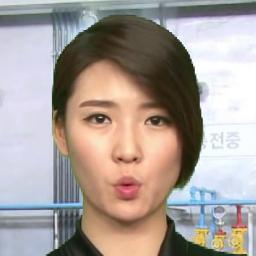}
			&\includegraphics[scale=0.25]{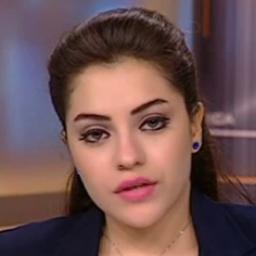}
			&\includegraphics[scale=0.25]{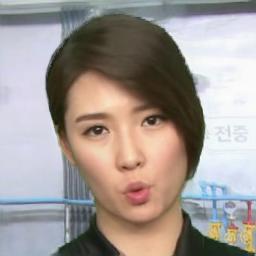}
			&\includegraphics[scale=0.25]{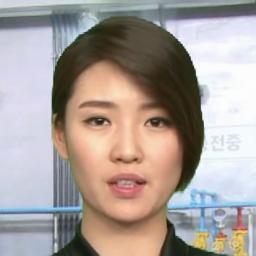}
			&\includegraphics[scale=0.25]{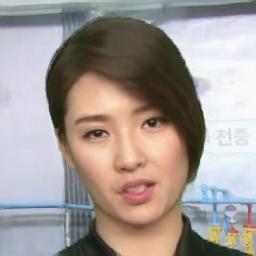}\\
			
			\includegraphics[scale=0.25]{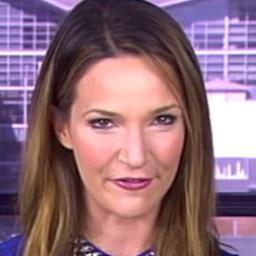}
			&\includegraphics[scale=0.25]{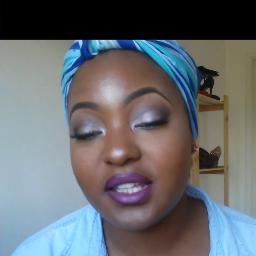}
			&\includegraphics[scale=0.25]{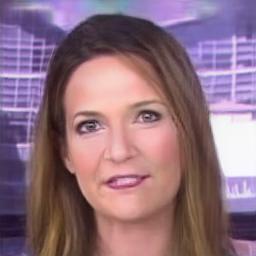}
			&\includegraphics[scale=0.25]{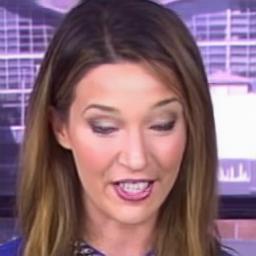}
			&\includegraphics[scale=0.25]{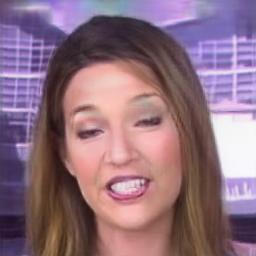}\\
			
			\includegraphics[scale=0.25]{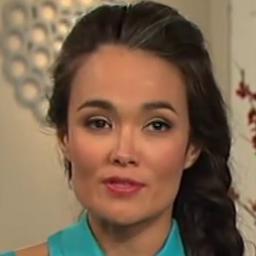}
			&\includegraphics[scale=0.25]{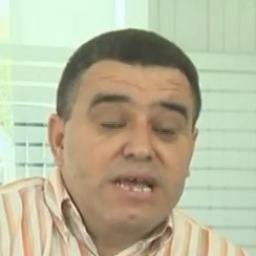}
			&\includegraphics[scale=0.25]{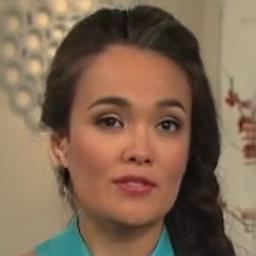}
			&\includegraphics[scale=0.25]{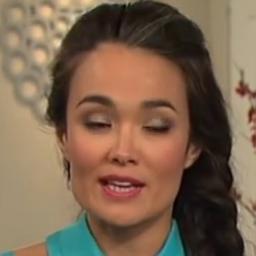}
			&\includegraphics[scale=0.25]{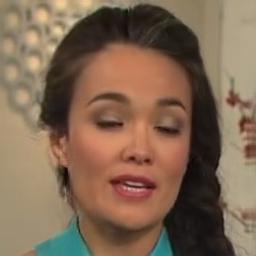}\\
			
			\includegraphics[scale=0.25]{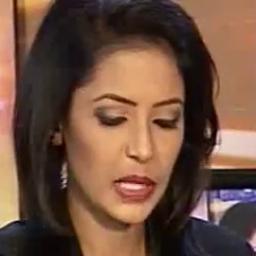}
			&\includegraphics[scale=0.25]{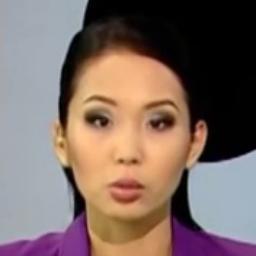}
			&\includegraphics[scale=0.25]{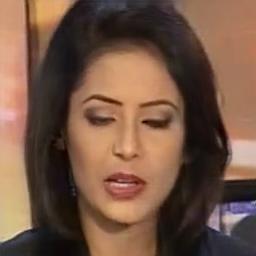}
			&\includegraphics[scale=0.25]{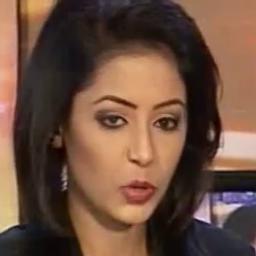}
			&\includegraphics[scale=0.25]{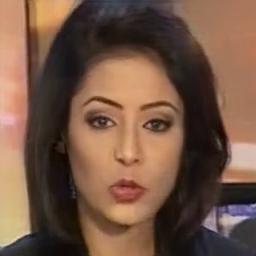}\\
			
			\includegraphics[scale=0.25]{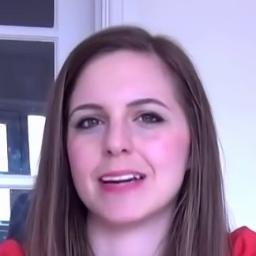}
			&\includegraphics[scale=0.25]{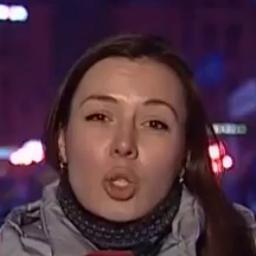}
			&\includegraphics[scale=0.25]{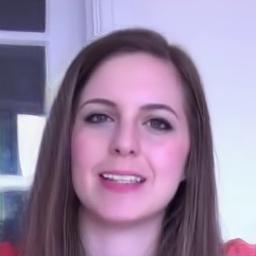}
			&\includegraphics[scale=0.25]{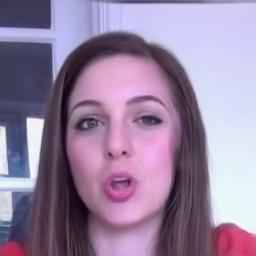}
			&\includegraphics[scale=0.25]{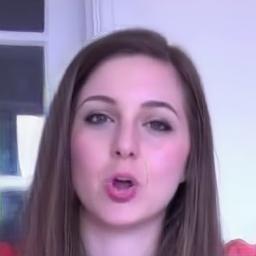}\\			
			$I_s$ &$I_d$ &\textbf{only pose} &\textbf{only expression} &\textbf{pose + expression}\\
	\end{tabular}}
	\caption{Result of pose-and-expression disentangle.}
	\label{fig:distangle}
\end{figure*}

\end{document}